\documentclass[lettersize,journal]{IEEEtran}
\usepackage{amsmath,amsfonts}
\usepackage{array}
\usepackage[caption=false,font=normalsize,labelfont=sf,textfont=sf]{subfig}
\usepackage{textcomp}
\usepackage{stfloats}
\usepackage{url}
\usepackage{verbatim}
\usepackage{graphicx}
\usepackage{cite}
\usepackage{graphicx}
\usepackage{multirow}
\usepackage{times}
\usepackage{amsmath,amsthm}
\usepackage{amssymb}
\usepackage{amsfonts}
\usepackage{color}
\usepackage{mathrsfs}
\usepackage{booktabs} 
\usepackage{makecell}
\usepackage{array}
\usepackage{caption}

\usepackage{color}
\usepackage{hyperref} 
\usepackage{url}
\usepackage{algpseudocode}
\usepackage{pifont}
\usepackage{bm}
\usepackage{comment}
\usepackage{wasysym}
\usepackage{bbding-zy}
\usepackage{float}
\usepackage{threeparttable}

\usepackage[utf8]{inputenc}

\usepackage[usenames,dvipsnames]{xcolor}
\usepackage{tikz} \usetikzlibrary{calc, arrows.meta, intersections, patterns, positioning, shapes.misc, fadings, through,decorations.pathreplacing}

\def\1{{\bf{1}}}
\def\0{{\bf{0}}}

\def\y{{\bf y}}

\def\H{{\bf H}}

\def\W{{\bf W}}

\def\Acal{{\mathcal{A}}}

\def\Ecal{{\mathcal{E}}}

\def\Gcal{{\mathcal{G}}}

\def\Lcal{{\mathcal{L}}}

\def\Ncal{{\mathcal{N}}}

\def\Rcal{{\mathcal{R}}}

\def\Vcal{{\mathcal{V}}}

\def\Rbb{{\mathbb R}}

\newtheorem{definition}{\textbf{Definition}}

\definecolor{ColorOne}{named}{MidnightBlue}
\definecolor{ColorTwo}{named}{Dandelion}
\definecolor{ColorThree}{named}{Plum}

\begin{document}

\title{Graph Dimension Attention Networks for
Enterprise Credit Assessment}


\author{Shaopeng~Wei,
        Béni~Egressy,
        Xingyan~Chen,
        Yu~Zhao,
        Fuzhen~Zhuang,~\IEEEmembership{Member,~IEEE},
        Roger~Wattenhofer,
        Gang~Kou
\thanks{S. Wei is with School of Business,  Guangxi University, China.
\protect\\
E-mail: shaopeng.wei@gxu.edu.cn}

\thanks{B. Egressy and R. Wattenhofer are with Information Technology and Electrical Engineering Department, ETH Zürich, Switzerland.
\protect\\
E-mail: begressy@ethz.ch}

\thanks{X. Chen and Y. Zhao are with Fintech Innovation Center, Financial Intelligence and Financial Engineering Key Laboratory of Sichuan Province, Institute of Digital Economy and Interdisciplinary Science Innovation, Southwestern University of Finance and Economics, China. 
\protect\\
E-mail: zhaoyu@swufe.edu.cn}

\thanks{F. Zhuang is with Institute of Artificial Intelligence, Beihang University, Beijing, China, and with Zhongguancun Laboratory, Beijing, China. 
Email:zhuangfuzhen@buaa.edu.cn}

\thanks{G. Kou is with School of Business Administration, Faculty of Business Administration, Southwestern University of Finance and Economics, China.}

\thanks{Y. Zhao and G. Kou (E-mail: kougang@swufe.edu.cn) are the corresponding authors.}

}


\markboth{Journal of \LaTeX\ Class Files,~Vol.~14, No.~8, August~2021}%
{Shell \MakeLowercase{\textit{et al.}}: A Sample Article Using IEEEtran.cls for IEEE Journals}

\IEEEpubid{0000--0000/00\$00.00~\copyright~2021 IEEE}

\maketitle

\begin{abstract}
Enterprise credit assessment is critical for evaluating financial risk, and Graph Neural Networks (GNNs), with their advanced capability to model inter-entity relationships, are a natural tool to get a deeper understanding of these financial networks.
However, existing GNN-based methodologies predominantly emphasize entity-level attention mechanisms for contagion risk aggregation, often overlooking the heterogeneous importance of different feature dimensions, thus falling short in adequately modeling credit risk levels.
To address this issue, we propose a novel architecture named \textbf{G}raph \textbf{D}imension \textbf{A}ttention \textbf{N}etwork (\textbf{GDAN}), which incorporates a dimension-level attention mechanism to capture fine-grained risk-related characteristics.
 Furthermore, we explore the interpretability of the GNN-based method in financial scenarios and propose a simple but effective data-centric explainer for GDAN, called \textbf{GDAN-DistShift}.
DistShift provides edge-level interpretability by quantifying distribution shifts during the message-passing process.
Moreover, we collected a real-world, multi-source \textbf{E}nterprise \textbf{C}redit \textbf{A}ssessment \textbf{D}ataset (\textbf{ECAD}) and have made it accessible to the research community since high-quality datasets are lacking in this field.
Extensive experiments conducted on ECAD demonstrate the effectiveness of our methods. In addition, we ran GDAN on the well-known datasets SMEsD and DBLP, also with excellent results.
\end{abstract}


\section{Introduction}
\label{sec:introduction}
\IEEEPARstart{E}{nterprise} credit assessment is a crucial process, essential for evaluating the credit risk of businesses. Timely identification of these risks provides a solid foundation for economic health and financial stability, helping to avoid considerable financial losses.
Graph Neural Networks (GNNs) possess advanced capabilities for modeling relationships between entities, making them a natural tool for comprehending financial networks and addressing the challenge of enterprise credit assessment.

\begin{figure}[t]
    \centering
    \includegraphics[width=0.45\textwidth]{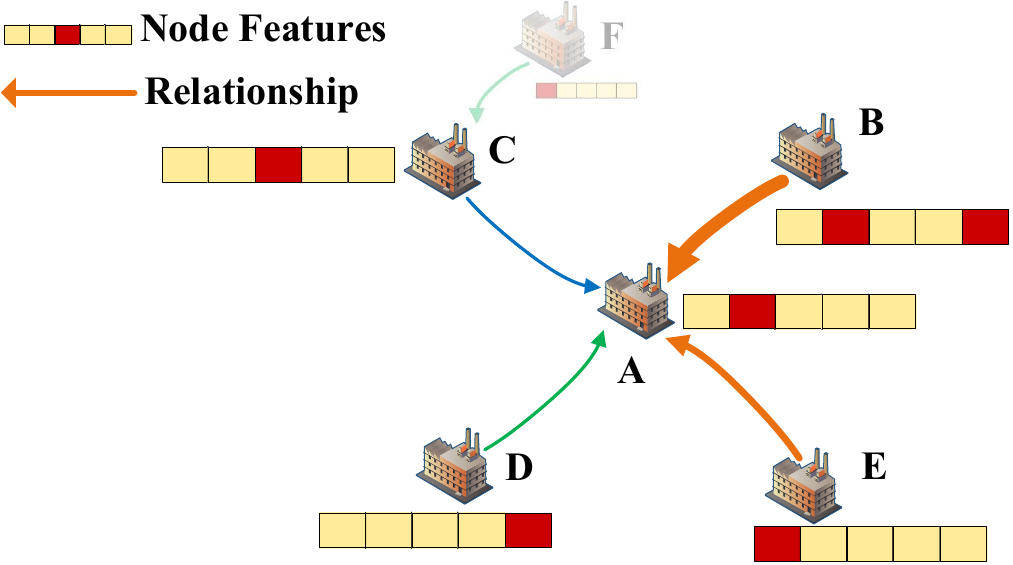}
    \caption{An Illustrative Example of Enterprise Heterogeneous Graph (EHG).
    The depth of color shading conveys the significance of each feature dimension, while the thickness of the edges reflects the relative importance of the respective relationships.}
    \vspace{-1em}
    \label{fig:scenario}
\end{figure}

\IEEEpubidadjcol

However, current GNN-based risk evaluation models focus on learning entity-level attention for nodes, often overlooking the fact that different feature dimensions encapsulate distinct aspects of an enterprise's risk profile. 
To illustrate, consider the case of Enterprise Heterogeneous Graphs (EHG) depicted in Figure \ref{fig:scenario}. 
Specifically, Enterprise A is connected to four neighbors—B, C, D, and E—through three types of relationships: customer (orange), investor (green), and supplier (blue). Each enterprise is characterized by five features: registered capital, time of establishment, number of legal disputes, administrative rewards, and administrative penalties.
Notably, Enterprise B is a recently established high-tech firm that heavily relies on government subsidies during its early stages. In this context, government penalties can significantly impact its future cash flow, thereby elevating its credit risk. These two critical risk factors—short establishment time and government penalties—are reflected by the second and fifth dimensions in deep red.
Enterprise C is part of a large supply chain and faces the risk of disrupted order quantities due to legal disputes with upstream or downstream enterprises, which is reflected in the third dimension. It is noted that, although Enterprise C is not as significant as Enterprise B at the entity level, it still possesses a critical feature that should not be under-weighted.
Conventional GNN-based methods assign entity-level weights to neighbors for the target enterprise (e.g., Enterprise A), thereby lacking the ability to discern differences among various dimensions of features, which is critical for capturing risk-related information. Consequently, these methods inadequately represent contagion risk, also known as risk spillover \cite{jegadeesh1993returns}.

To address this crucial issue, we introduce \textbf{G}raph \textbf{D}imension \textbf{A}ttention \textbf{N}etworks (\textbf{GDAN}) as a novel solution to comprehensively model enterprise credit assessment (see Definition \ref{definition-enterprise credit assessment}) within EHG (see Definition \ref{definition-Enterprise heterogeneous graph}). In particular, GDAN leverages a dimension-level attention mechanism to autonomously ascertain the significance of individual risk dimensions, thereby enabling the model to discern subtle distinctions among different risk aspects of neighboring nodes. Furthermore, we consider the specific type of each relationship to enhance the modeling of risk-related information.

\begin{definition}
\label{definition-Enterprise heterogeneous graph}
\textbf{Enterprise Heterogeneous Graph}. 
An enterprise heterogeneous graph (EHG) is a graph $\Gcal=(\Vcal,\Ecal, \Acal, \Rcal, \H)$, where $\Vcal$ denotes the set of all nodes, $\Ecal$ denotes the set of edges, $\Acal$ denotes the set of node types, $\Rcal$ denotes the set of edge types and $\H$ denotes the enterprise feature matrix (node input features) consisting of business, lawsuit, and other risk information. An EHG has two associated mapping functions: (i) an node mapping function $\Phi$: $\Vcal \to \Acal$, where $\forall v_i \in \Vcal$, $\Phi(v_i) \in \Acal$, $ |\Acal| \geq 1$.
(ii) an edge mapping function $\phi$: $\Ecal \to \Rcal$, where $\forall e_i \in \Ecal$, $ \phi(e_i) \in \Rcal$, $|\Rcal| \geq 2$. 
\end{definition}

\begin{definition}
\label{definition-enterprise credit assessment}
\textbf{Enterprise Credit Assessment (ECA)}. 
Given an enterprise heterogeneous graph, which consists of enterprise risk information and multiple relationship types among enterprises, we aim to assess the enterprise credit level of each node. This can be treated as a multi-class node classification problem.
\end{definition}

Another significant issue pertains to the model interpretability, which is a pronounced concern in financial domain.
While GDAN is built on deep neural networks, rendering it black-box systems for decision-makers.
In the context of GNNs, interpretability hinges on the message-passing process and the complex network of interactions.
Specifically, GDAN can change the distribution of input features through the message passing process, which we define as \textbf{dist}ribution \textbf{shift}, to make them easier to distinguish. 
Therefore, we propose a simple and effective explainer, termed \textbf{GDAN-DistShift}, to provide data-centric edge-level interpretability for GDAN. 
To enhance the transparency of DistShift, we employ a simple machine learning (ML) model as a proxy for measuring distribution in tandem with a straightforward aggregation technique.

Furthermore, the accessibility of data presents a notable impediment to advancing research within the financial domain. 
The research contributions of studies such as \cite{yang2020financial,wang2021temporal,bi2022company,zheng2023midlg} have inspired advancements in this field. However, it is regrettable that the datasets used in these works are not openly available.
Consequently, we collect a comprehensive \textbf{E}nterprise \textbf{C}redit \textbf{A}ssessment \textbf{D}ataset (\textbf{ECAD}), which we release for public use.\footnote{The codebase and datasets required for replication purposes are accessible on GitHub: \url{https://github.com/shaopengw/GDAN}.} The \textbf{ECAD} encompasses detailed information about 48,758 small and medium-sized enterprises (SMEs), encompassing fundamental business data, litigation records, and records of administrative licenses, rewards and penalties, as well as over one million relationships between them.
Finally, we conduct extensive experiments on both the ECAD dataset and publicly available datasets, SMEsD and DBLP, to validate the efficacy of our proposed GDAN model in capturing critical risk information and to assess the effectiveness of DistShift for interpretability.

The contributions of this work are threefold:

\begin{itemize}
    \item We propose \textbf{G}raph \textbf{D}imension \textbf{A}ttention \textbf{N}etwork (\textbf{GDAN}), a novel GNN model.
    \textbf{GDAN} leverages the dimension attention mechanism within the message-passing process, enabling it to discern and capture pivotal risk-related information.
    To the best of our knowledge, this is the first work to propose dimension-level attention for enterprise credit assessment.
    \item Furthermore, we present \textbf{GDAN-DistShift}, a simple and effective data-centric interpretability tool for GDAN. 
    DistShift evaluates edge importance as interpretability by capturing the changes in feature distributions caused by message passing along edges. This approach offers greater transparency compared to previous deep neural network-based GNN interpretability methods. 
    \item To alleviate the problem of the lack of high-quality, open-access datasets in finance, we manually collect a multi-source dataset, \textbf{ECAD}, for enterprise credit assessment and make it available to the broader research community. We conduct comprehensive experiments utilizing the \textbf{ECAD} dataset, SMEsD dataset, and the widely-recognized DBLP dataset, which demonstrates the effectiveness of \textbf{GDAN} and \textbf{DistShift}.
\end{itemize}

\section{Related Work}
\subsection{Financial Risk Assessment}
Financial risk encompasses the potential for an enterprise to encounter financial difficulties, default on payments or declare bankruptcy. It is often quantified by credit levels, a comprehensive indicator taking multiple factors into account. 
Most previous studies in this area primarily center on financial metrics and econometric methodologies \cite{lo1986logit,crouhy2000comparative}. For example, Lopez and Saidenberg \cite{lopez2000evaluating} introduce a credit assessment approach involving cross-sectional simulations based on panel data spanning multiple years and various assets. Contemporary investigations have integrated machine learning and deep learning techniques, employing methods such as Support Vector Machines (SVM), decision trees, and deep neural networks \cite{olson2012comparative,delen2013measuring,tsai2008using}. For instance, 
Zhang et al. \cite{zhang2022credit} employ deep neural networks to amalgamate demographic and behavioral data, enhancing the modeling of credit risk for SMEs.
Recent research has also begun to investigate the impact of multi-modal data on financial issues, incorporating sources such as enterprise annual reports, conference calls, and social media.
For example, Craja et al. \cite{craja2020deep} utilize a hierarchical attention network to extract risk-related information from structured documents, such as annual reports, and integrate this with financial ratios to detect instances of financial statement fraud. Borchert et al. \cite{borchert2023extending} take advantage of textual website content and apply deep neural networks to forecast business failures.
Chen et al. \cite{chen2023bankruptcy} find that annual report text-based communicative value effectively mitigates the model misidentification of a non-bankrupt firm as a bankrupt firm.
li et al. \cite{li2024corporate} quantify climate risk exposure at the firm level through textual analysis of earnings call transcripts.

However, these methods are all based on sufficient financial indicators, which are often not available for SMEs. Moreover, most of them do not consider the complex interaction among entities in a financial market, limiting their performance on EHGs.

\subsection{Graph Neural Networks}
Graph Neural Networks (GNNs) aim to utilize the topological information inherent in network structures. This structure can originate from real-world data sources \cite{Rong2020Self,hu2020heterogeneous} or be constructed from unstructured data, including images and text \cite{huang2022contexting,xv2023commerce}. For example, Hu et al. \cite{hu2020heterogeneous} integrate transformer-based attention mechanisms into GNNs for a huge academic network. Xu et al. \cite{Xu2017Scene} generate scene graphs from original input images and acquire structured scene representations through GNNs. 
Early GNNs predominantly focused on homogeneous graphs, such as Graph Convolution Networks (GCN) \cite{Kipf2017Semi-supervised}, wherein target nodes aggregate information equally from neighboring nodes. 
Tsitsulin et al \cite{tsitsulin2023graph} find that GNN pooling methods doesn't work well at clustering graphs. Therefore, they introduce Deep Modularity Networks (DMoN), an unsupervised pooling method for clustering structure of real-world graphs.
Many recent GNN applications are based on heterogeneous graphs, which can have different node and relationship types.
For example, Wang et al. \cite{Wang2019Heterogeneous} apply a hierarchical attention mechanism to model recommendation systems, considering multiple relationships between users and items. 
From the perspective of technique, applying attention mechanism is one of the most popular way for GNNs \cite{fountoulakis2023graph}.
Graph Attention Networks (GAT) \cite{Velickovic2018Graph} introduced an attention mechanism to enable GNNs to autonomously learn weights for neighboring nodes. 
Lee et al. \cite{lee2023towards} propose AEROGNN, aiming to mitigate over-smoothed features and smooth cumulative attention related to deep graph attention.

The application of GNNs in financial contexts is natural due to the wealth of structural information present in financial markets. For instance, Yang et al. \cite{yang2021financial} propose a dynamic graph-based GNN to model the default risk of supply chain enterprises. 
Zheng et al. \cite{zheng2021heterogeneous} propose triple-level attention-based GNNs to predict the bankruptcy of SMEs.
Xiang et al. \cite{xiang2023semi} leverage transaction records to construct temporal transaction graphs and use GNNs to find fraud patterns.
Shi et al. \cite{shi2024improved} first construct edges using kNN and then utilize GNNs to predict credit risk. Their findings show significant improvements compared to using only internal information.
Wei et al. \cite{wei2024combining} also leverage GNNs to combine enterprise intra-risk and inter-risk for bankruptcy prediction.


Nevertheless, it is worth noting that existing GNN-based methods primarily employ entity-level attention, potentially overlooking critical risk factors within specific dimensions.

\subsection{Interpretability of GNNs}
Interpretability has garnered significant attention in the era of deep learning \cite{nagarajan2019uniform,angelov2020towards}. 
It is critical, particularly in domains such as finance, where individuals require a comprehensive understanding of model causality to make confident, informed, and explainable decisions.
Liu et al. \cite{liu2021mining} focus on both the performance and interpretability of credit risk assessment. They propose an automatic feature-crossing method to generate cross features for Logistic Regression (LR), thereby endowing the final model with improved performance and interpretability.
Dumitrescu et al. \cite{dumitrescu2022machine} employ decision trees to augment the capabilities of LR, with the dual objective of capturing non-linear effects while preserving intrinsic interpretability. 
It is noteworthy that the quest for interpretability in the context of GNNs necessitates a unique consideration of interactions among data points. Specifically, optimal interpretability should elucidate the impact of edges, pinpointing the crucial edges that contribute to a model's decision.
For example, Ying et al. \cite{ying2019gnnexplainer} propose GNNExplainer, a classical model designed to identify important features, nodes, and subgraphs for GNN interpretability. GNNExplainer achieves this by maximizing the mutual information between the prediction of the original graph and learned patterns.
Yuan et al. \cite{yuan2021explainability} develop SubgraphX, which aims to identify important subgraphs through Monte Carlo tree search.
Yin et al. \cite{yin2023train} refer to t pre-training techniques and propose to distill the universal
interpretability of GNNs by pre-training over synthetic graphs with ground-truth explanations.

However, most existing research primarily focuses on node representation, with less attention given to modeling the edges.
Furthermore, current GNN explanation methods predominantly emphasize model behavior rather than graph data. These methods often employ deep learning techniques to identify significant patterns in GNNs, which may introduce opacity into the interpretability process and potentially obscure the intrinsic nature of GNNs.

\section{Methodology}
\subsection{Graph Dimension Attention Networks}
\begin{figure*}[htb]
    \centering
    \includegraphics[width=0.99\textwidth]{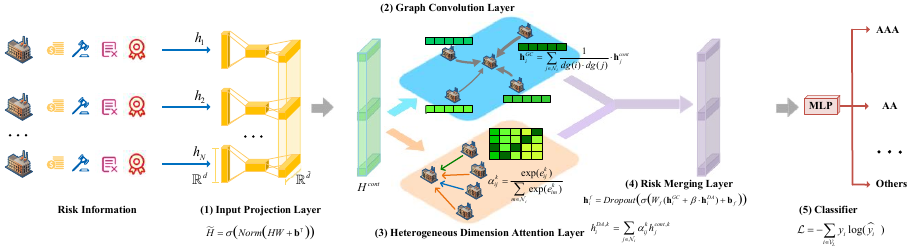}
    \caption{Model Architecture of GDAN. (1) Input Projection Layer aims to map original risk features into a shared latent space; (2) Graph Convolution Layer learns structural information of EHG; (3) Heterogeneous Dimension Attention Layer captures risk-related information regarding different dimensions; (4) Risk Merging Layer fuse former risk information to generate final risk representations; (5) Classifier gives the final assessed risk level. }
    \label{fig:GDAN}
\end{figure*}

This section formally defines GDAN. As illustrated in Figure \ref{fig:GDAN}, the GDAN model consists of five parts: (1) Input Projection Layer; (2) Graph Convolution Layer; (3) Heterogeneous Dimension Attention Layer; (4) Risk Merging Layer; (5) Classifier and Optimization.

\subsubsection{Input Projection Layer}
The Input Projection Layer maps input node features into a common vector space whilst aiming to extract important features.
Given an input feature matrix $\H \in \Rbb^{N \times d}$, where $N$ denotes the number of enterprises and $d$ denotes the number of risk dimensions. The input projection layer transforms it into the $\Rbb^{N \times \Tilde{d}}$ latent space:
\begin{equation}
\begin{array}{l}
\begin{aligned}
    \label{input_projection_layer}
     \Tilde{\H}=\sigma \big( \textit{Norm} \big( \H \W + \textbf{b}^\intercal \big ) \big ) \ ,
\end{aligned}
\end{array}
\end{equation}
where $\W \in \Rbb^{d \times \Tilde{d}}$ and $\textbf{b} \in \Rbb^{\Tilde{d}}$ are trainable parameters, $\textit{Norm}$ denotes batch normalization, and $\sigma$ is the Relu activation function. 

For general heterogeneous graphs with multiple node types and node type mapping function $\Phi: \Phi(v_i) \to \Acal$, 
and for all $a \in \Acal$, $\Vcal_{a}:= \{v \in \Vcal \mid \Phi(v) = a \} \subseteq \Vcal$ 
and $H_a \in \Rbb^{|{\Vcal_{a}}| \times d_a}$,
we project nodes of different types into the same latent space:

\begin{equation}
\begin{array}{l}
\begin{aligned}
    \label{input_projection_layer_multi-node_type}
    \H_a^{\prime} =\sigma \big( \H_a \W_a  + \textbf{b}_a^\intercal \big  ) \ ,
\end{aligned}
\end{array}
\end{equation}
where $\W_a \in \Rbb^{d_a \times \Tilde{d}}$ and $\textbf{b}_a \in \Rbb^{\Tilde{d}}$ are trainable parameters and $\sigma$ is the Relu activation function. We then implement Eq. (\ref{input_projection_layer}) based on the concatenation of all types of nodes representations $\H^{\prime}$.

\subsubsection{Graph Convolution Layer}
We generate contagion risk representations for each node as follows:
\begin{equation}
\begin{array}{l}
\begin{aligned}
    \label{node_encoding_layer}
    \textbf{h}_i^{cont} = \W_{cont} \Tilde{\textbf{h}}_i + \textbf{b}_{cont} \ ,
\end{aligned}
\end{array}
\end{equation}
where $\W_{cont}$ and $\textbf{b}_{cont}$ are trainable parameters. $\textbf{h}_i^{cont}$ denotes learned contagion risk.
To learn the natural structure information, we conduct basic graph convolution as follows:
\begin{equation}
\begin{array}{l}
\begin{aligned}
    \label{graph_convolution_layer}
    \textbf{h}_i^{GC} = \sum_{j \in \Ncal_{i}} \frac{1}{\textit{dg}(i) \cdot \textit{dg}(j)} \cdot \textbf{h}_j^{cont} \ ,
\end{aligned}
\end{array}
\end{equation}
where $dg(\cdot)$ represents the degree function,  $\Ncal_i$ denotes neighbors of node $i$. $\textbf{h}_j^{cont}$ and $\textbf{h}_i^{GC}$ are the contagion representation of neighbor node $j$  and the Graph-Convolution risk representation of node $i$, respectively.

\subsubsection{Heterogeneous Dimension Attention Layer}
Afterwards, we augment the risk representation by enhancing specific dimensions automatically, enabling the model to capture risk-related information across different dimensions.
Moreover, to sufficiently model the heterogeneous relations in EHG, we transform heterogeneous neighbor information according to the type of the relation: $e_{ij}= \textbf{h}_j^{cont} \textbf{W}_{r} $, where $j \in \Ncal_{i}^{r} $ and
$\textbf{W}_{r}$ denotes the transformation matrix for $r \in \Rcal$ .

Then we calculate the attention score of dimension $k$ regarding target node $i$ and its neighbors of all types. Specifically, we use the $\textit{Softmax}$ function to normalize the dimension importance across all neighbors that target node $i$ has as follows:
\begin{equation}
\begin{array}{l}
\begin{aligned}
    \label{hete_dimension_attention_layer-weight}
     \alpha_{ij}^k  & = \textit{Softmax}(e_{ij}^{k})  \\
                  & = \frac{\exp(e_{ij}^{k})}{\sum_{m \in \Ncal_{i}} \exp(e_{im}^{k})} \ ,
\end{aligned}
\end{array}
\end{equation}
where $\alpha_{ij}^k$ is the learned importance of dimension $k$  in terms of target node $i$ and neighbor node $j$. $m \in \Ncal_i$ denotes one neighbor node of $i$.
We aggregate the contagion risk for each node based on the learned attention score of dimensions. Each dimension of Dimension-attention-based risk representation is calculated as follows:
\begin{equation}
\begin{array}{l}
\begin{aligned}
    \label{hete_dimension_attention_layer-aggregation}
       h_i^{DA,k} = \sum_{j \in \Ncal_{i}} \alpha_{ij}^k h_j^{cont,k}.
\end{aligned}
\end{array}
\end{equation}

\subsubsection{Risk Merging Layer}
We merge the Dimension-Attention-based and Graph-Convolution-based risk representations to generate comprehensive risk representations. We utilize a hyper-parameter $\beta$ to balance the two representations as follows:
\begin{equation}
\begin{array}{l}
\begin{aligned}
    \label{risk_merging_output_layer-merge}
       \textbf{h}_i^{f} = \textit{Dropout} \big( \sigma \big( \W_f (\textbf{h}_i^{GC}+\beta \cdot \textbf{h}_i^{DA}) + \textbf{b}_f \big) \big) \ ,
\end{aligned}
\end{array}
\end{equation}
where $\textbf{h}_i^f$ denotes the enhanced risk representation of node $i$. $\W_f$ and $\textbf{b}_f$ are trainable parameters. $\sigma$ is the activation function; we use Relu in our models. We use dropout to alleviate potential problems with over-fitting. 

\subsubsection{Classifier and Optimization}
Then, we utilize an MLP layer to generate final risk representations of the target node and normalize it using $\textit{Softmax}$:
\begin{equation}
\begin{array}{l}
\begin{aligned}
    \label{risk_merging_output_layer-output}
       \hat{y_i}^c = \textit{Softmax} \big( \textit{MLP}\big( \textbf{h}_i^f\big) \big) \ ,
\end{aligned}
\end{array}
\end{equation}
where $\hat{y_i}^c$ denotes the probability that node $i$ belongs to risk level $c$. During inference, the risk level with the highest probability $\hat{y_i}$, is taken as the prediction.

The enterprise credit assessment can be treated as a node classification problem, thus we train our model by minimizing the cross-entropy loss as follows:
\begin{equation}
\begin{array}{l}
\begin{aligned}
    \label{optimization}
       \Lcal=-\sum \limits_{i \in \Vcal_L}  y_i\log(\hat{y_i})\ ,
\end{aligned}
\end{array}
\end{equation}
where $\Vcal_{L}$ is the set of enterprises with ground truth credit levels.  $y_i$ is the one-hot encoding of the ground truth label for node $i$ and $\hat{y_i}$ is the vector of predicted probabilities for each credit level for node $i$.

\subsection{Discussion of Dimension Attention}
\begin{figure}[t]
    \centering
    \includegraphics[width=\linewidth]{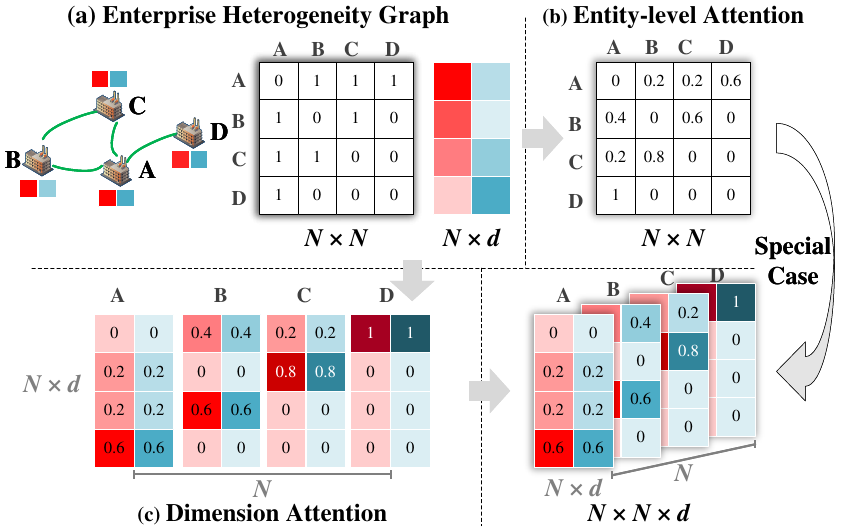}
    \caption{Illustration of the difference between entity-level attention and dimension attention.}
    \vspace{-1em}
    \label{fig:motivation}
\end{figure}

\textbf{Comparing with Entity Level Attention.}
As illustrated in Figure  \ref{fig:motivation} part (a), an EHG comprises an adjacency matrix $A \in \Rbb^{N \times N}$ and node features $H \in \Rbb^{N \times d}$.
Conventional GNNs with entity-level attention generate attention scores, creating an 
$N \times N$ matrix, as shown in part (b). However, these models overlook variations in importance across different feature dimensions.
In GNNs with dimension attention, 
attention scores are learned by accounting for the contributions of various feature dimensions. This results in an $N \times d$ matrix for each node, as depicted in part (c). When all dimensions contribute equally, the dimension attention scores degenerate into entity-level attention scores.
We believe that the dimension attention mechanism empowers GNN models to capture more nuanced information, thereby enhancing their ability to better fit the data.

\textbf{Time Complexity Analysis.}
We confirm that it is feasible to apply the approach to large datasets because the time complexity of our method correlates linearly with the size of the graphs.
In our approach, we perform three crucial steps in the dimension attention operation, detailed in $e_{ij}= \textbf{h}_j^{cont} \textbf{W}_{r} $, Eq. (\ref{hete_dimension_attention_layer-weight}) and Eq. (\ref{hete_dimension_attention_layer-aggregation}). To break it down further, when we consider the edge count $|\mathcal{E}|$ and node count $|\mathcal{V}|$ to represent the graph's scale, we find that:

\begin{itemize}
    \item The time complexity of the neighbor-relation based operation, $e_{ij}= \textbf{h}j^{cont} \textbf{W}{r}$, is $\mathcal{O}(|\mathcal{E}|)$.
    \item For the equation $\alpha_{ij}^k = \textit{Softmax}(e_{ij}^{k}) =\frac{\exp(e_{ij}^{k})}{\sum_{m \in \mathcal{N}_{i}} \exp(e_{im}^{k})} $, the denominator's time complexity is $\mathcal{O}(|\mathcal{V}|)$ due to node summarization. This operation can be done with parallel computing in GPU, which makes it more efficient in practice. Meanwhile, the numerator's count of $e_{ij}^{k}$ scales linearly with $|\mathcal{E}|$.
    \item Furthermore, the summarization operation in Eq. (\ref{hete_dimension_attention_layer-aggregation}) scales linearly with $|\mathcal{V}|$.
\end{itemize}

Taking all these factors into account, the total time complexity sums up to $\mathcal{O}(|\mathcal{V}|+|\mathcal{E}|)$. This finding assures the scalability of our proposed model when dealing with large datasets.

\subsection{DistShift for GDAN}
In the realm of financial decision-making, interpretability is as crucial as performance itself. To address this need, we propose GDAN-DistShift, a simple yet effective data-centric explanation method for GDAN.

GNNs have demonstrated superior efficacy compared to earlier machine learning and deep learning approaches when applied to graph data. 
This heightened performance can be attributed to GNNs' unique capacity to leverage both the feature information of data points and the intricate interactions among nodes.

Given an input that includes the original features of nodes, $\H$, it is feasible to employ various techniques to extract salient features and subsequently perform classification, as follows:
\begin{equation}
\begin{array}{l}
\begin{aligned}
    \label{distshift_ml-dl}
       \hat{y}_0 = \textbf{Classifier} \big (f(\H) \big) \ ,
\end{aligned}
\end{array}
\end{equation}

\begin{figure}[t]
    \centering
    \includegraphics[width=0.47\textwidth]{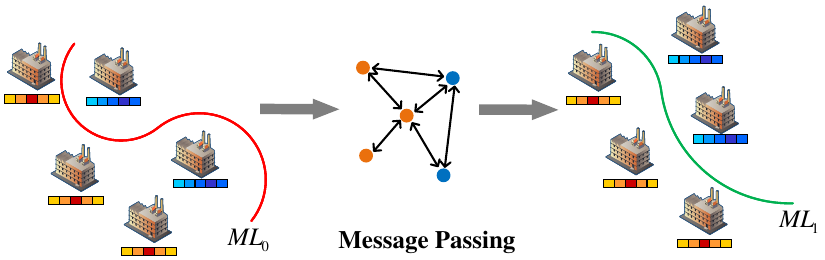}
    \caption{Distribution Shift. Following the message passing procedure intrinsic to GNNs, the revised node attributes exhibit enhanced distinguishability.}
    \vspace{-1em}
    \label{fig:distshift}
\end{figure}

The reason we use GNNs is that there are edges that help to improve the model features of input nodes. In essence, all GNN-based classification models can be simplified as follows:
\begin{equation}
\begin{array}{l}
\begin{aligned}
    \label{distshift_mpgnn}
       \hat{y}_1 = \textbf{Classifier} \big (\textbf{GNN}(\H,\Gcal) \big) \ ,
\end{aligned}
\end{array}
\end{equation}
When comparing Equation (\ref{distshift_mpgnn}) and Equation (\ref{distshift_ml-dl}), a notable observation emerges: alterations in the input characteristics of classifiers yield consequential shifts in the embeddings, which can ultimately lead to changes in model performance. Consequently, as Figure \ref{fig:distshift} shows, when employing a common ML model like LR as the classifier, models founded on GNNs can perform better than models relying solely on the traditional ML paradigm.

We therefore introduce a novel method to assess edge importance by quantifying the distribution shift during message passing. We believe that this approach is effective for GDAN, as it is a typical message-passing based GNN model.
In an ideal scenario with ground truth labels for all the nodes in a dataset, we would compare the probability for the true class with an edge and without it. If the presence of the edge increases the true class prediction, then we would consider this an important edge. However, in EHG scenario, the ground truth labels are sparse. So we instead utilize the entropy of the predicted probabilities as a proxy. The idea is that low entropy outputs are easier to separate in the embedding space, and therefore edges that decrease entropy for a node will be considered important.


To enhance stability, we employ an unbalanced sampling strategy for subgraph selection, thereby comprehensively evaluating edge significance across a range of subgraphs.

\subsubsection{Unbalanced Sampling}
The ECAD dataset exhibits a pronounced imbalance characterized by uneven distributions of node degrees, node types, and edge types. This substantial imbalance poses the risk of compromising the performance of GNNs.
To mitigate this issue, we introduce weighted assignments to nodes and edges, based on their respective node degrees, node types, and edge types. 
Within the ECAD dataset, shareholder relationships constitute a predominant portion of all edge types. This preponderance can potentially impede the efficacy of GNNs operating on heterogeneous graphs, as such models may struggle to discern between distinct relationship types. Consequently, we opt to diminish the weights associated with shareholder relationships.
Similarly, we augment the weights assigned to node types that represent a smaller proportion among all node types. This adjustment aims to enable the model to effectively distinguish between various node types. 
Specifically, we set the probability of node type $c$ as $1-2 \cdot \frac{|\Vcal_c|}{|\Vcal|}$, where $|\Vcal_c|$ and $|\Vcal|$ denote the number of nodes that belong to node type $c$ and all nodes, respectively.
Furthermore, we establish the sampling ratio of neighbor nodes in accordance with their node degrees, wherein nodes with higher degrees are afforded a greater likelihood of being sampled. This strategic allocation is based on the premise that nodes with higher degrees often wield more influence within their respective communities.

The sampling process commences with a set of initial seed nodes denoted as $b$. Subsequently, their neighbors are sampled at a rate defined by probability $p$, which is determined by the product of the corresponding probabilities associated with node type, edge type, and node degree. Note that the sampling is done across all neighbor nodes at once, rather than sampling individually for each target node.
The neighbors that are sampled alongside the initial seed nodes are then designated as the new seeds for the subsequent iteration. This process repeated $l$ times. 
The sampled subgraphs, in conjunction with the entire graph, constitute the graph set $\Gcal_{sub}$ for later calculation. 
Note that including the entire graph ensures that all edges will be assigned an importance.

\subsubsection{Distribution Shift}
The key idea revolves around capturing the distribution shift following message passing iterations, specifically gauging the extent to which an edge contributes to facilitating the classification of a target node. 

We adopt a machine learning model as a quantifiable metric for assessing the distribution of features.
Given an input feature matrix $\H$, we evaluate nodes' credit levels with a  machine learning model $\textbf{ML}_0$ as $\hat{\y}_0 = \textbf{ML}_0(\H)$.
We use the entropy of the predicted probabilities of node $i$, $\textit{entropy}(i)$ as its distance to the center of the distribution learned by $\textbf{ML}_0$. 
We are looking for a low entropy, which indicates a high confidence, or rather a high separation from the decision boundary. 


We refer to the message-passing process in GCN, where each node averages its neighbors' information. For transparency, we use a simple average aggregation function $AGG$ instead of more complex aggregation schemes to simulate the message-passing process:
\begin{equation}
\begin{array}{l}
\begin{aligned}
    \label{distshift_aggregation}
       \Tilde{h}_i = \sum_{j \in \Ncal_i} \frac{1}{\sqrt{dg(j)} \sqrt{dg(i)}} h_j + h_i,
\end{aligned}
\end{array}
\end{equation}


Eq. (\ref{distshift_aggregation}) is actually a non-linear version of GCN. Prior research \cite{chen2020revisiting,he2020lightgcn,wu2019simplifying} has shown through experiments and theory that GCNs utilizing linear aggregation can achieve comparable performance to non-linear GCNs. 

Then we train a new ML model based on the updated feature matrix $\Tilde{\H}$ as $\hat{\y}_1 = \textbf{ML}_1(\Tilde{\H})$.
In this setting, each edge can change the target node by 
delivering a source node's information and we can evaluate such influence.
We assume deleting or adding one edge doesn't change the distribution of all features. 
For each edge $e_{ij}$ that connects target node $i$ and source node $j$, it updates the target node $i$'s information as follows:
\begin{equation}
\begin{array}{l}
\begin{aligned}
    \label{distshift_message_passing}
       \tilde{h_{ij}} =  \frac{1}{\sqrt{dg(j)} \sqrt{dg(i)}} \Tilde{h}_j + \Tilde{h}_i,
\end{aligned}
\end{array}
\end{equation}
We then compare the entropy of predicted probabilities regarding target node $i$ of 
$\textbf{ML}_1\big(AGG\big(\H_{i,j}, E=\{e_{ij}\}\big)\big)$ and $\textbf{ML}_0\big(AGG\big(\H_{i,j}, E=\emptyset \big)\big)$ to calculate the importance of edge $e_{ij}$. 

\section{Experiments}
\begin{table}[t]
\centering
\caption{Statistics of Datasets}
\label{tab:dataset}
\resizebox{\linewidth}{!}{
\begin{tabular}{c|c|l|c|ll}
\toprule
\multicolumn{3}{c}{\textbf{Datasets}}          & \multicolumn{1}{|l}{\textbf{Training}} & \textbf{Validation} & \textbf{Testing} \\ \midrule 
\multirow{7}{*}{\textbf{ECAD}}  & \multicolumn{1}{c}{\textbf{\#Node}}  &                 & \multicolumn{3}{c}{48758}                           \\ \cline{2-6} 
                      & \multirow{3}{*}{\textbf{\#Relation}} & \textbf{\#\textit{Investor\_holder}} & \multicolumn{3}{c}{77572}                           \\
                      &                           & \textbf{\#\textit{Branch} }        & \multicolumn{3}{c}{12764}                           \\
                      &                           & \textbf{\# \textit{Shareholder}}    & \multicolumn{3}{c}{1302415}                         \\  \cline{2-6}
                      & \multirow{3}{*}{\textbf{\#Label}}    & \textbf{\#AAA}           & \multicolumn{1}{c}{327}      & \multicolumn{1}{c}{122}        & \multicolumn{1}{c}{117}     \\
                      &                           & \textbf{\#AA}         & \multicolumn{1}{c}{378}      & \multicolumn{1}{c}{108}        & \multicolumn{1}{c}{91}      \\
                      &                           & \textbf{\#Others}            & \multicolumn{1}{c}{74}       & \multicolumn{1}{c}{28}         & \multicolumn{1}{c}{29}      \\ \midrule
\multirow{3}{*}{\textbf{SMEsD}} & \multicolumn{2}{c|}{\textbf{\#Node}}                   & \multicolumn{3}{c}{3976}                           \\ \cline{2-6}
                      & \multicolumn{2}{c|}{\textbf{\#Relation}}               & \multicolumn{3}{c}{31170}                          \\ \cline{2-6}
                      & \multicolumn{2}{c|}{\textbf{\#Label}}                  & \multicolumn{1}{c}{2816}    & \multicolumn{1}{c}{721}        & \multicolumn{1}{c}{491}  \\   \midrule
\multirow{3}{*}{\textbf{DBLP}} & \multicolumn{2}{c|}{\textbf{\#Node}}                   & \multicolumn{3}{c}{26128}                           \\ \cline{2-6}
                      & \multicolumn{2}{c|}{\textbf{\#Relation}}               & \multicolumn{3}{c}{239566}                          \\ \cline{2-6}
                      & \multicolumn{2}{c|}{\textbf{\#Label}}                  & \multicolumn{1}{c}{3245}    & \multicolumn{1}{c}{406}        & \multicolumn{1}{c}{477}  \\

                      \bottomrule  
\end{tabular}}
\vspace{-1em}
\end{table}

\begin{table*}[t]
\begin{center}
\caption{Enterprise Credit Assessment and Node Classification Results. The top two scores in each column are marked in bold and underlined.}
\vspace{-1em}
\label{tab:main_results}
\renewcommand{\arraystretch}{1.3}
\resizebox{\linewidth}{!}{
\begin{tabular}{cccccccccccc}
\toprule
\multicolumn{1}{l}{} & \multicolumn{4}{c}{ECAD}                                                                       & \multicolumn{3}{c}{SMEsD}                                                            & \multicolumn{3}{c}{DBLP}                                        \\ \toprule
\multicolumn{1}{l}{} &         & ACC                 & F1                  & \multicolumn{1}{c|}{AUC}                 & ACC                 & F1                  & \multicolumn{1}{c|}{AUC}                 & ACC                 & F1                  & AUC                 \\ \hline
\multirow{3}{*}[-3pt]{ML}  & LR      & 57.81               & 40.73               & \multicolumn{1}{c|}{64.84}               & 74.95               & 55.6                & \multicolumn{1}{c|}{80.69}               & 79.45               & 79.02               & 93.79               \\ \cline{2-11} 
                     & SVM     & 60.76               & 42.17               & \multicolumn{1}{c|}{65.67}               & 65.38               & 10.53               & \multicolumn{1}{c|}{52.21}               & 80.08               & 79.39               & 94.16               \\ \cline{2-11} 
                     & GBDT    & 61.60                & 54.98               & \multicolumn{1}{c|}{77.82}               & 74.13               & 47.74               & \multicolumn{1}{c|}{83.00}                  & 77.36               & 76.95               & 92.89               \\ \hline
\multirow{7}{*}{GNNs} & GCN     & 66.67±0.60          & 47.89±0.98          & \multicolumn{1}{c|}{80.46±1.99}          & 76.21±0.42          & 71.91±0.75          & \multicolumn{1}{c|}{75.72±2.01}          & 90.31±0.97          & 90.05±1.08          & 98.20±0.06          \\ \cline{2-11} 
                     & GAT     & 62.53±1.40          & 44.56±1.90          & \multicolumn{1}{c|}{80.64±0.50}          & 76.41±1.04          & 71.81±1.90          & \multicolumn{1}{c|}{78.35±1.60}          & 91.86±0.88          & 91.60±0.94          & 98.32±0.38          \\ \cline{2-11} 
                     & HAN     & 63.61±0.46          & 49.27±6.47          & \multicolumn{1}{c|}{80.29±0.57}          & 76.65±0.33          & 72.52±0.45          & \multicolumn{1}{c|}{76.48±1.67}          & 93.46±1.19          & 93.19±1.27          & 98.99±0.32          \\ \cline{2-11} 
                     & HGT     & 66.87±0.70          & 63.42±0.71          & \multicolumn{1}{c|}{78.71±1.04}          & 76.57±0.45          & 73.46±0.33          & \multicolumn{1}{c|}{80.73±0.60}          & 93.42±0.75          & 93.22±0.75          & 98.86±0.30          \\ \cline{2-11} 
                     & SeHGNN  & 68.10±1.05          & \underline{65.58±2.46}          & \multicolumn{1}{c|}{\underline{82.10±0.33}}          & 71.03±0.24          & 66.27±0.37          & \multicolumn{1}{c|}{72.89±0.84}          & \underline{95.15±0.48}          & \underline{94.91±0.50}          & \underline{99.15±0.11}          \\ \cline{2-11} 
                     & HAT     & 64.83±0.98          & 57.63±4.35          & \multicolumn{1}{c|}{74.02±1.57}          & \underline{76.78±0.72}          & 72.83±1.45          & \multicolumn{1}{c|}{78.11±2.24}          & 85.32±1.45          & 84.86±1.35          & 96.40±0.40          \\ \cline{2-11}
\multicolumn{1}{l}{} & ComRisk & \underline{68.57±0.88}          & 64.07±6.32          & \multicolumn{1}{c|}{78.63±3.63}          & 76.45±0.76          & \underline{73.67±0.98}          & \multicolumn{1}{c|}{\underline{80.76±1.01}}          & 92.55±0.81          & 92.24±0.93          & 98.12±0.23          \\ \hline
\multicolumn{2}{c}{GDAN (Ours)}       & \textbf{72.39±1.54} & \textbf{69.00±1.81} & \multicolumn{1}{c|}{\textbf{84.51±0.72}} & \textbf{78.81±0.46} & \textbf{75.30±0.94} & \multicolumn{1}{c|}{\textbf{84.84±0.66}} & \textbf{95.51±0.48} & \textbf{95.33±0.53} & \textbf{99.36±0.09} \\ \bottomrule
\end{tabular}}
\end{center}
\vspace{-1em}
\end{table*}

\subsection{Datasets}
\textbf{ECAD.} Due to the lack of high quality public financial risk datasets, we construct a new dataset \textbf{ECAD}, for enterprise credit assessment, by collecting data from multiple sources.
We first sample a subset of 512 enterprises, from the population of enterprises with credit rating records. We then search for their one-hop neighbors according to three relationship types, which are detailed below. Afterwards, we crawl public risk information for those enterprises. The final dataset consists of 48,758 enterprises, including their multi-source risk information and more than one million connections between them. There are 1274 enterprises that have labels, including good, normal and bad, which refer to AAA, AA and others (i.e., A, BBB, BB etc.), respectively.
The risk information includes an enterprise's basic business information, lawsuit information, administrative licenses, rewards and penalties.
We preprocess the raw risk information to transform it into numbers as features.
There are three types of relationships among enterprises: \textit{Investor\_holder} relations, \textit{Branch} relations and \textit{Shareholder} relations.
We split the enterprises with labels into train, validation, and test sets according to a $(0.6, 0.2, 0.2)$ ratio. We show the statistics of the dataset in Table \ref{tab:dataset}.




\textbf{SMEsD.} To further validate the performance of GDAN, we also implement experiments on another financial risk dataset, SMEsD. SMEsD consists of enterprises and 
 heterogeneous relationships among them, and is used for bankruptcy prediction. We adhere to the dataset settings in the paper \cite{wei2024combining}.

\textbf{DBLP.} To demonstrate the effectiveness of the proposed model, we also conduct experiments on the common academic dataset, DBLP, which is a heterogeneous graph with multiple node types and relations. We downloaded the dataset from Pytorch-geometric (PyG \cite{fey2019fast}) and split the author nodes into train, validation, and test sets as shown in Table \ref{tab:dataset}.

\subsection{Baselines}
We compare the proposed model with two types of current state-of-the-art (SOTA) baselines: (1) \textbf{Machine learning (ML) models}, including LR \cite{hosmer2013applied}, SVM \cite{suykens1999least} and GBDT \cite{friedman2001greedy}, which only consider the feature information of enterprise nodes or authors nodes; (2) Graph Neural Networks, which take the whole EHG graph into consideration. Specifically, the GNNs baselines can be further divided into \textbf{homogeneous graph-based GNNs} (GCN \cite{Kipf2017Semi-supervised}, GAT \cite{Velickovic2018Graph}), \textbf{heterogeneous graph-based GNNs} (HAN \cite{Wang2019Heterogeneous}, HGT \cite{hu2020heterogeneous} and SeHGNN \cite{yang2023simple}) and \textbf{GNNs for financial risk prediction} (HAT \cite{zheng2021heterogeneous} and ComRisk \cite{wei2024combining}). 
More details are shown as following:

\textit{Machine Learning Models:}
\begin{itemize}
    \item Logistic Regression (\textbf{LR}) \cite{hosmer2013applied}: This is a widely-used simple and efficient classification method.
    \item Support Vector Machine (\textbf{SVM}) \cite{suykens1999least}: This model utilizes support vectors to divide vector spaces into different classes.
    \item Gradient Boosting Decision Tree (\textbf{GBDT}) \cite{friedman2001greedy}: This is a classic tree classification model of conventional ML.
\end{itemize}

\textit{Graph Neural Networks}
\begin{itemize}
    \item Graph Convolutional Networks (\textbf{GCN}) \cite{Kipf2017Semi-supervised}: This model updates node embeddings iteratively by averaging neighboring representations.
    
    \item Graph Attention Networks (\textbf{GAT}) \cite{Velickovic2018Graph}: This model employs an attention mechanism to automatically assign different weights to neighbors during the message-passing process. 

    
    \item Heterogeneous Graph Attention Network (\textbf{HAN})
    \cite{Wang2019Heterogeneous}: This model proposes to learn heterogeneous graphs with a hierarchical attention mechanism.

    \item Heterogeneous Graph Transformer (\textbf{HGT} \cite{hu2020heterogeneous}): This model applies the transformer attention mechanism in the message-passing process on heterogeneous graphs.

    \item Simple and Efficient Heterogeneous Graph Neural Network (\textbf{SeHGNN} \cite{yang2023simple}): This model highlights the use of multi-hop metapath-based neighbor information using an attention mechanism.

   \item Heterogeneous-attention-network-based model (\textbf{HAT}) \cite{zheng2021heterogeneous}: This model leverages triple-level attention for SMEs' bankruptcy prediction.

    \item Graph Neural Networks for Enterprise
Bankruptcy Prediction (\textbf{ComRisk} \cite{wei2024combining}): This model utilizes hierarchical heterogeneous attention mechanism for enterprise bankruptcy prediction.
\end{itemize}

\begin{figure*}[htb]
    \centering
    \includegraphics[width=0.24\textwidth]{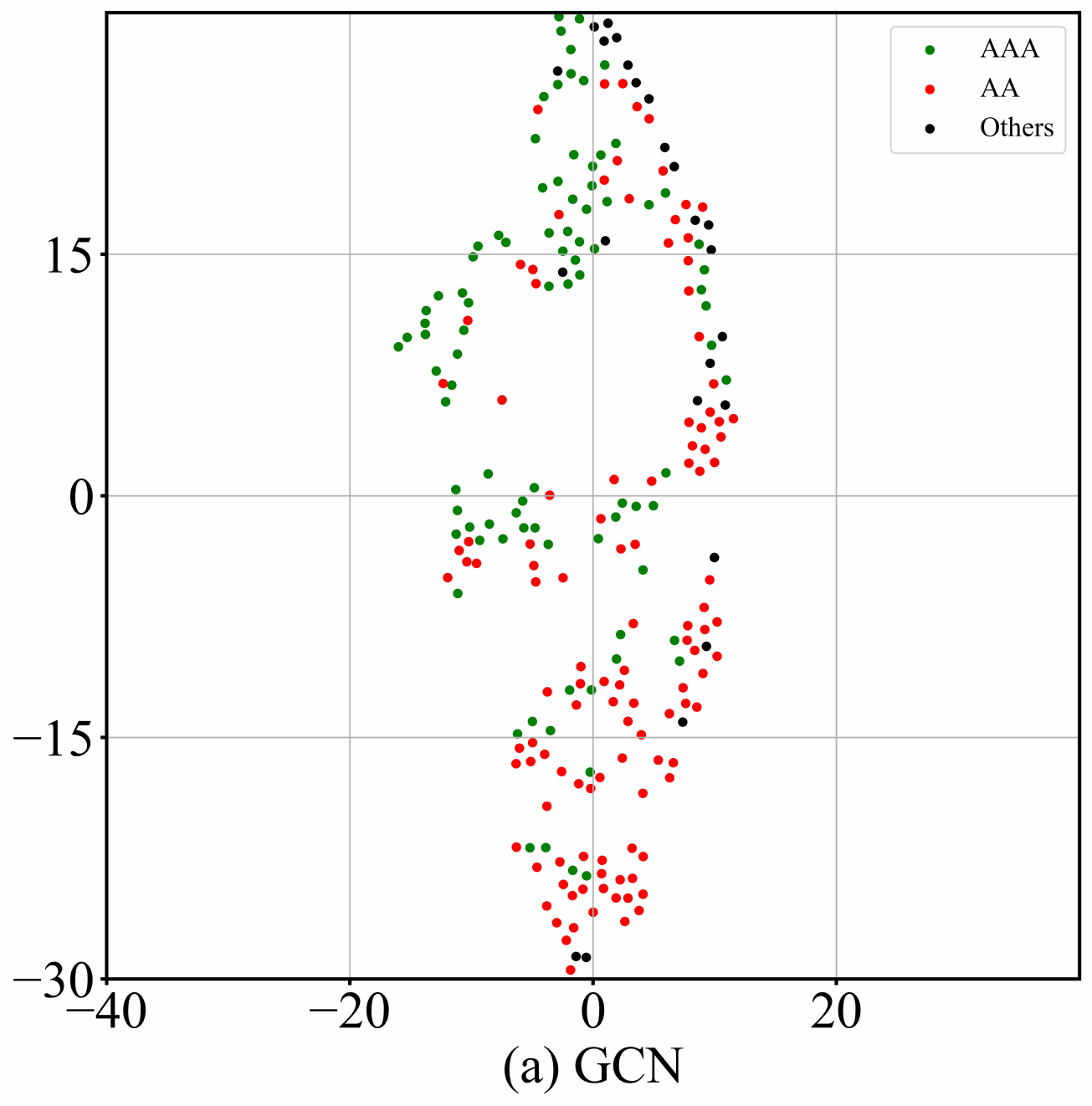}
    \includegraphics[width=0.24\textwidth]{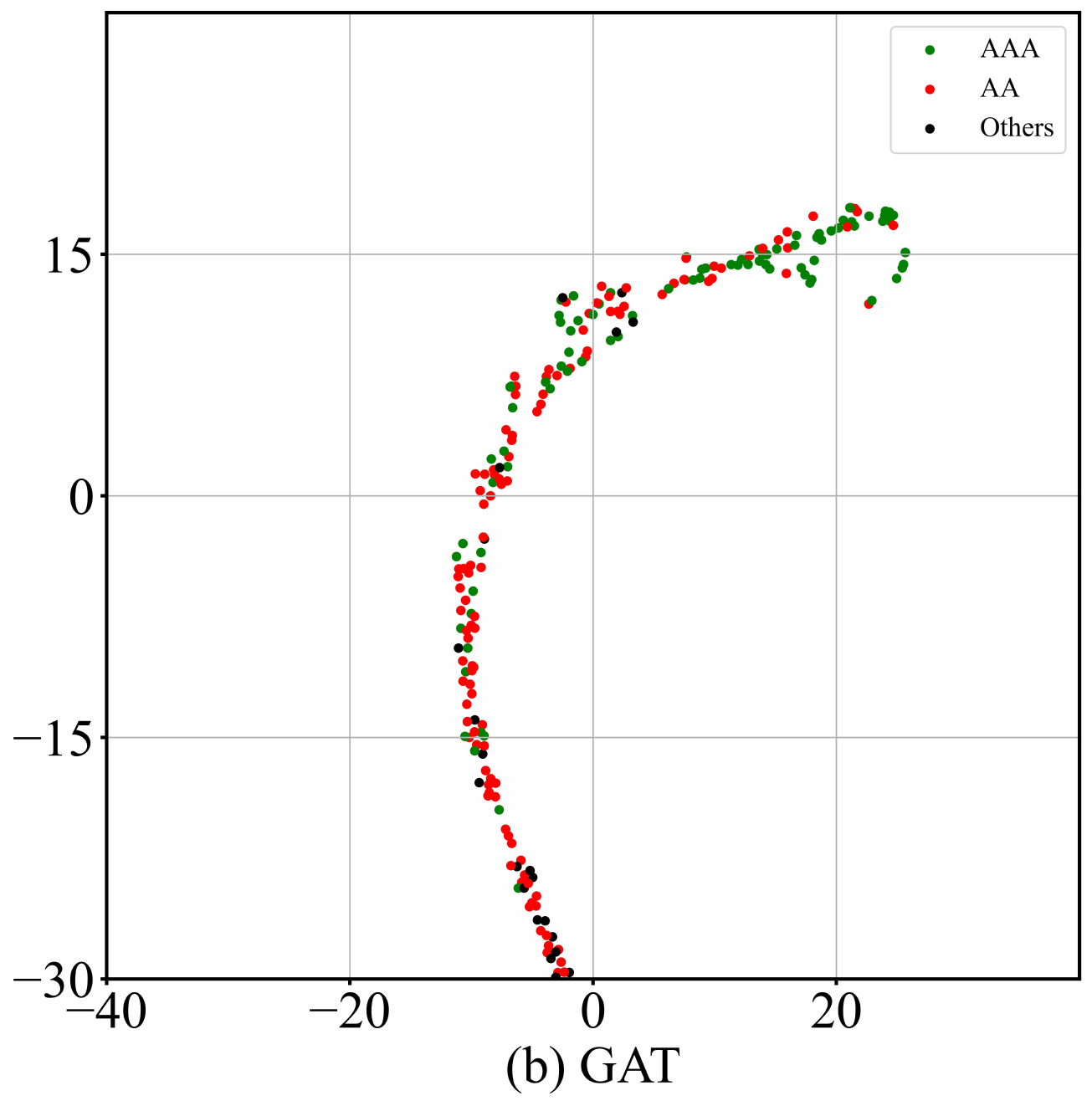}
    \includegraphics[width=0.24\textwidth]{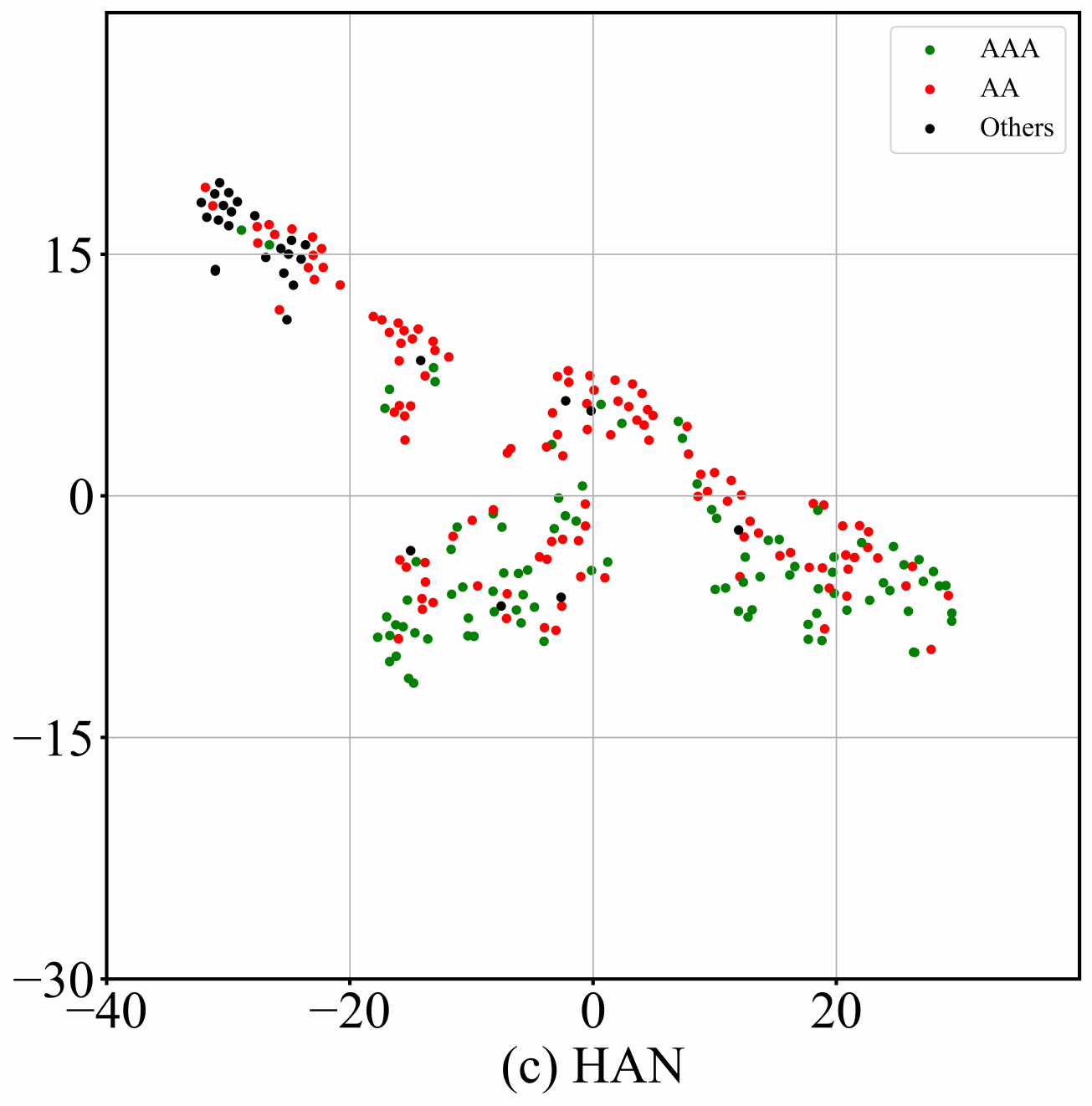}
    \includegraphics[width=0.24\textwidth]{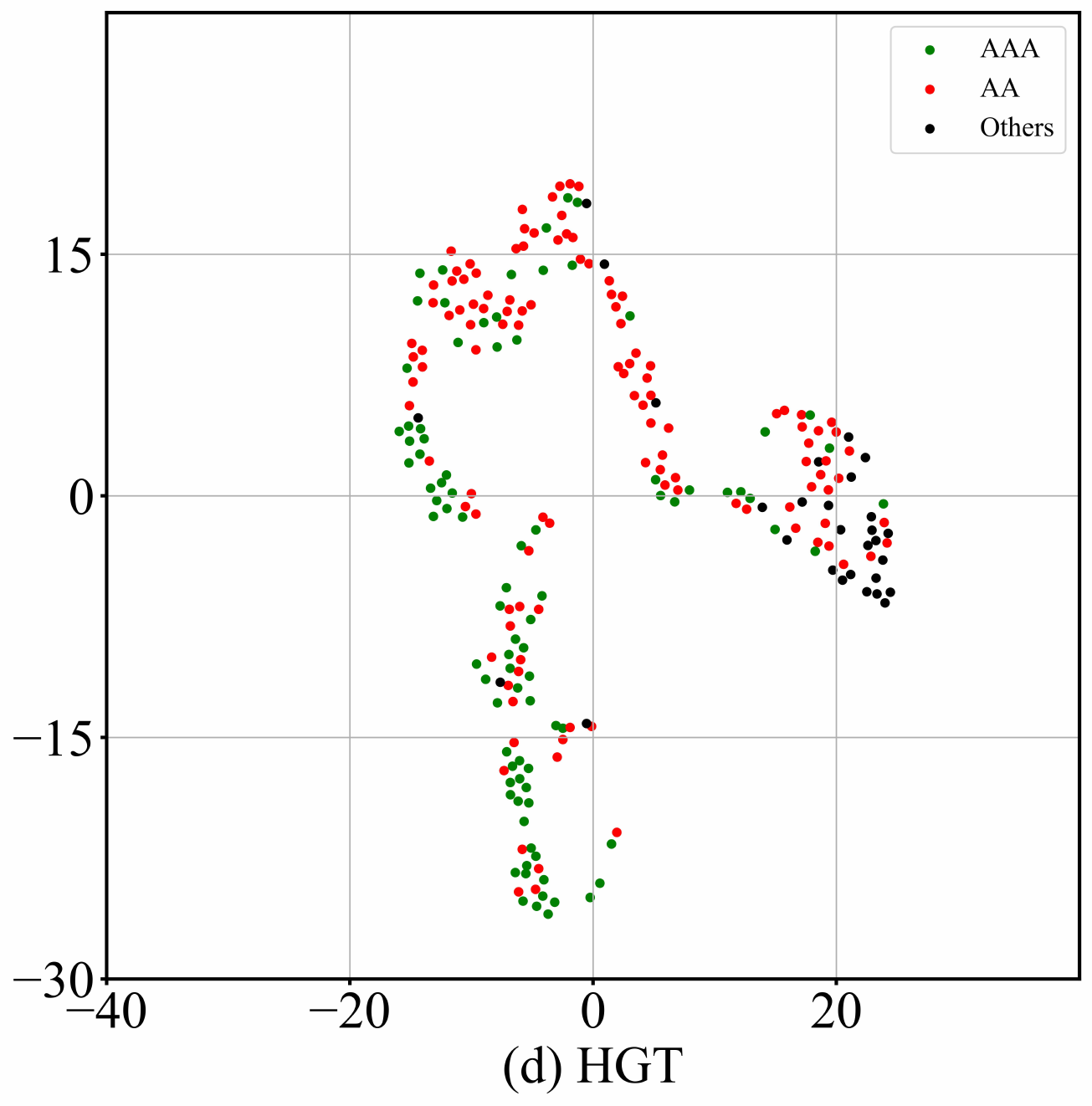}\\
    \includegraphics[width=0.24\textwidth]{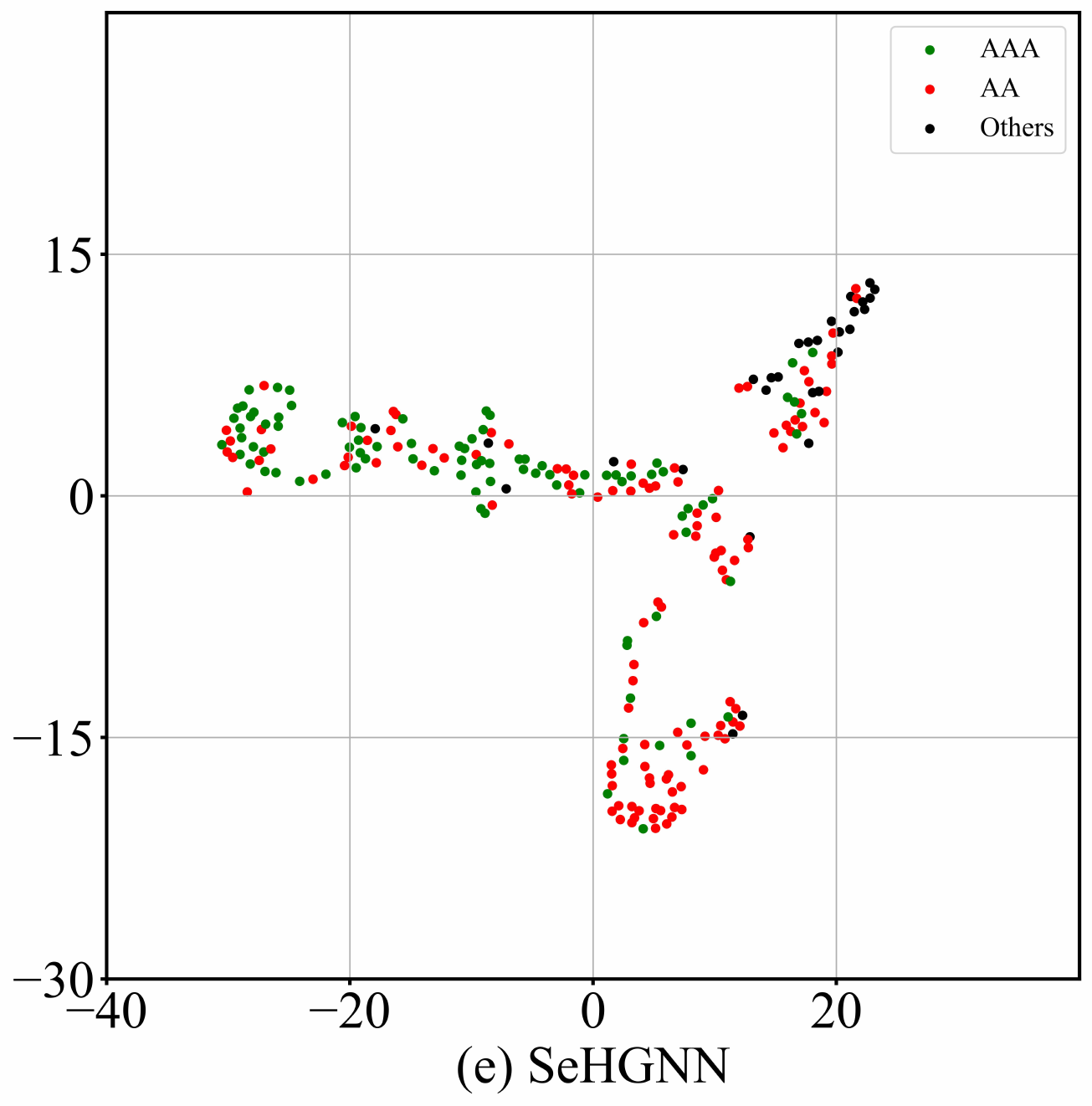}
    \includegraphics[width=0.24\textwidth]{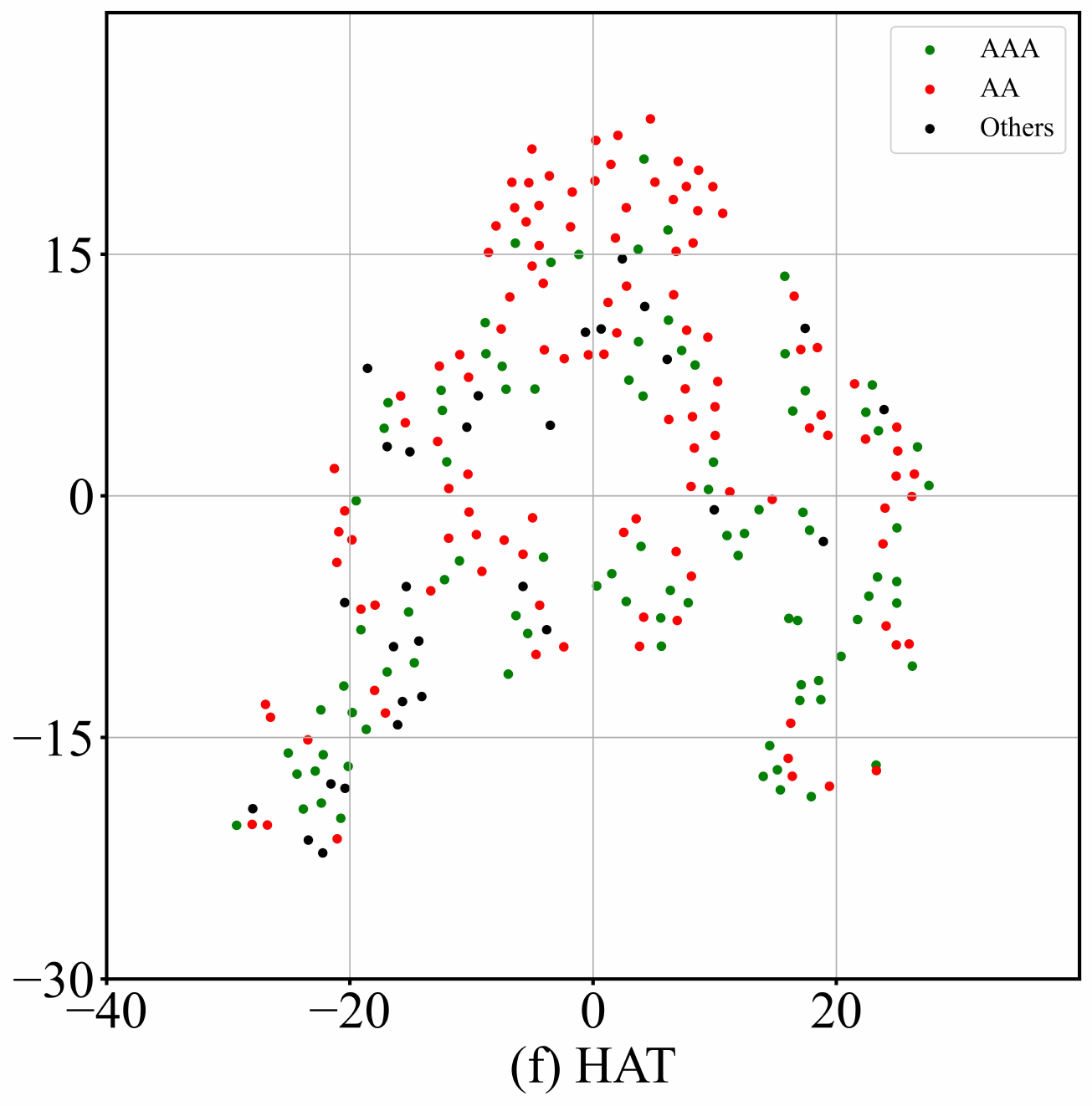}
    \includegraphics[width=0.24\textwidth]{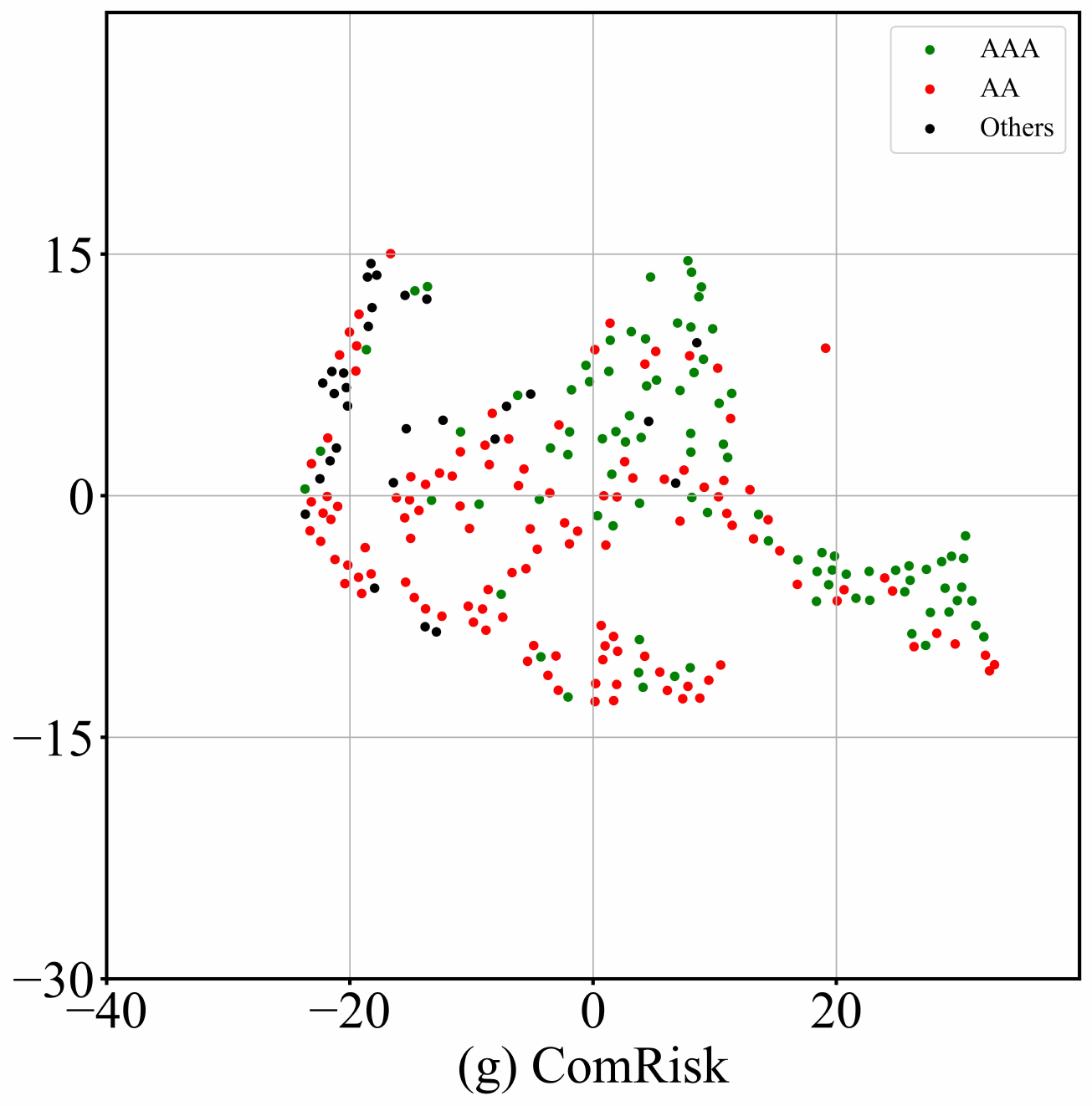}
    \includegraphics[width=0.24\textwidth]{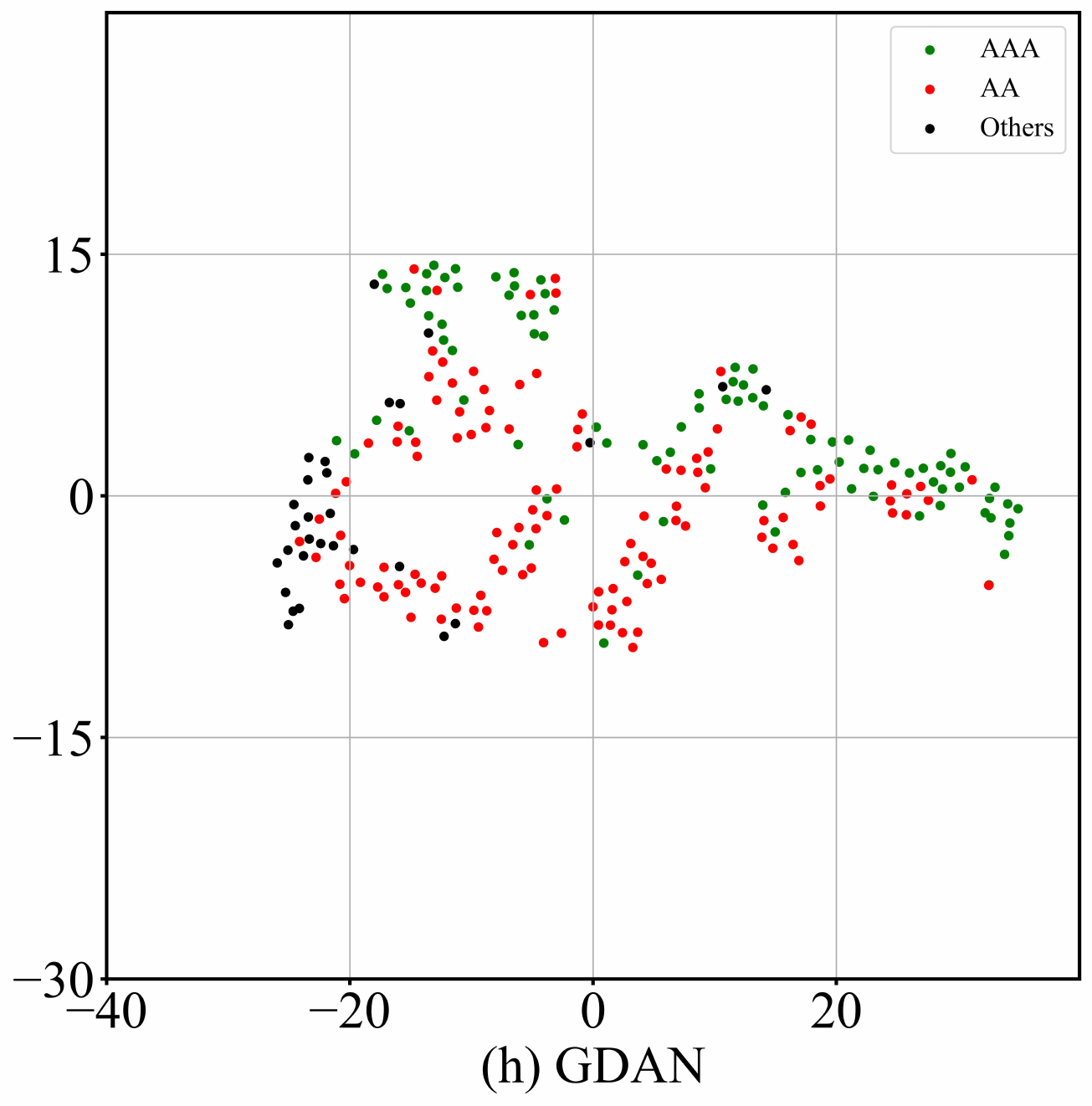}
    \caption{Visualization of node representations for different GNN models on ECAD. Green points, red points, and black points denote enterprises with ``AAA'', ``AA'' and ``Others'' credit levels, respectively. We can observe that GDAN outperforms baseline models. }
    \vspace{-1em}
    \label{fig:Visualization}
\end{figure*}

\subsection{Experiment Setting}
\label{sec-experiment_settings}
We implement the proposed model and all baseline models on three datasets (i.e., ECAD and DBLP) and evaluate them with three indicators (i.e., accuracy, macro-F1, and AUC). 
We use the scikit-learn package to run machine learning models. For SeHGNN, we use the official code in the paper \cite{yang2023simple} and we use the implementation code provided by PyG for all other GNN models. 
We run all the models five times and report the average performance. All the GNNs models are based on PyTorch and PyG and trained with AdamW \cite{loshchilov2017decoupled}. The learning rate is adjusted between 0.001 to 0.01 by the Cosine Annealing Learning Rate Scheduler \cite{loshchilov2016sgdr}. For the ECAD dataset, we set the input dimension, hidden dimension and output dimension (before classifier) for all GNN models as 99, 64, and 24, respectively. 
For the SMEsD dataset, we set the input dimension, hidden dimension and output dimension as 23, 18 and 12, respectively.
For DBLP, we fine-tune GNN models with dimensions from \{512, 256, 128, 64, 32\} for the hidden layers and the output layer (before classifier).
We run for 200 epochs on ECAD, 400 epochs on SMEsD, and for 100 epochs on DBLP. We choose the model with the best performance in terms of accuracy for ECAD and SMEsD, and in terms of macro-F1 for DBLP on the validation set and test it on the test set. 
We fine-tune all GNN models with layer numbers from \{1,2,4,8\}, head number from \{2,4,8\} and weight decay from \{0.1, 0.01, 0.001\}.

\subsection{ECA and Node Classification}

We report the experimental results of the proposed model and all baseline models in Table \ref{tab:main_results}.
We can observe that GDAN outperforms all baseline models on all three datasets. Specifically, our model achieves improvements of 3.82\% accuracy, 3.42\% macro-F1, and 2.41\% AUC against the state-of-art baseline models (ComRisk and SeHGNN) on ECAD. Our model also improves upon ComRisk, HAT and SeHGNN performances on the SMEsD and DBLP dataset by all measures. 
The results demonstrate the effectiveness of GDAN in learning representations on complex heterogeneous graphs.

\textbf{Analysis}. (1) We observe from Table \ref{tab:main_results} that the three ML models (i.e., LR, SVM and GBDT) achieve commendable results on the ECAD dataset, which confirms that the collected multi-source risk information enhances the assessment of a company's credit risk. 
Our model, when compared to these machine learning models, exhibits substantial improvements across all metrics on all datasets, confirming GDAN's ability to model topological information.
(2) Our investigation reveals that all GNN models outperform traditional ML models, underscoring the pivotal role of topology information. 
Furthermore, among both heterogeneous graph-based models (RGCN, HAN, HGT, SeHGNN) and GNNs for financial risk prediction (HAT and ComRisk), most of them outperform homogeneous graph-based models (GAT) on the ECAD dataset, underscoring the need to model heterogeneous relationships.
However, it is worth noting that GCN performs exceptionally well on the ECAD dataset, likely due to the predominance of edges of the same type, which can lead to over-fitting challenges for heterogeneous GNNs when encoding edge-type information in EHG. 
(3) In contrast, our model consistently outperforms all other graph neural network models on all datasets, emphasizing the significance of distinguishing between feature dimensions in the message-passing process.

\subsection{Node Clustering}
We employ node clustering to demonstrate GDAN's efficacy.
To begin, we execute a feed-forward process across all GNN models on the test set of ECAD to obtain the enterprise representations. Subsequently, we apply K-Means clustering to these representations to derive node clustering labels. 
We then assess the quality of the clustering results by comparing the predicted labels with the ground truth labels of the enterprises, utilizing normalized mutual information (NMI) and adjusted Rand index (ARI) as evaluation metrics. 
In order to enhance result reliability, we repeat this entire process 10 times and present the average results in Table \ref{tab:clustering}.
The experimental outcomes substantiate the superior performance of our model when compared with all the baseline models. 


\begin{table}[t]
\caption{Node Clustering Results}
\label{tab:clustering}
\resizebox{\linewidth}{!}{
\begin{tabular}{ccccccccc}
\toprule
\textbf{} & \multicolumn{1}{c}{\textbf{GCN}} & \multicolumn{1}{c}{\textbf{GAT}} & \multicolumn{1}{c}{\textbf{HAN}} & \multicolumn{1}{c}{\textbf{HGT}} & \multicolumn{1}{c}{\textbf{SeHGNN}} & \multicolumn{1}{c}{\textbf{HAT}} & \multicolumn{1}{c}{\textbf{ComRisk}} & \textbf{GDAN} \\ 
    \midrule
ARI & 0.0606                  & 0.0582                  & 0.0261                  & 0.0391                  & 0.0388                     & 0.0504                  & 0.0505                      & \textbf{0.0957} \\ \midrule
NMI & 0.1446                  & 0.1417                  & 0.1157                  & 0.1481                  & 0.1494                     & 0.0801                  & 0.1586                      & \textbf{0.1624} \\ \bottomrule
\end{tabular}}
\end{table}

 \begin{figure*}[t]    
    \centering
    \includegraphics[width=0.99\linewidth]
    {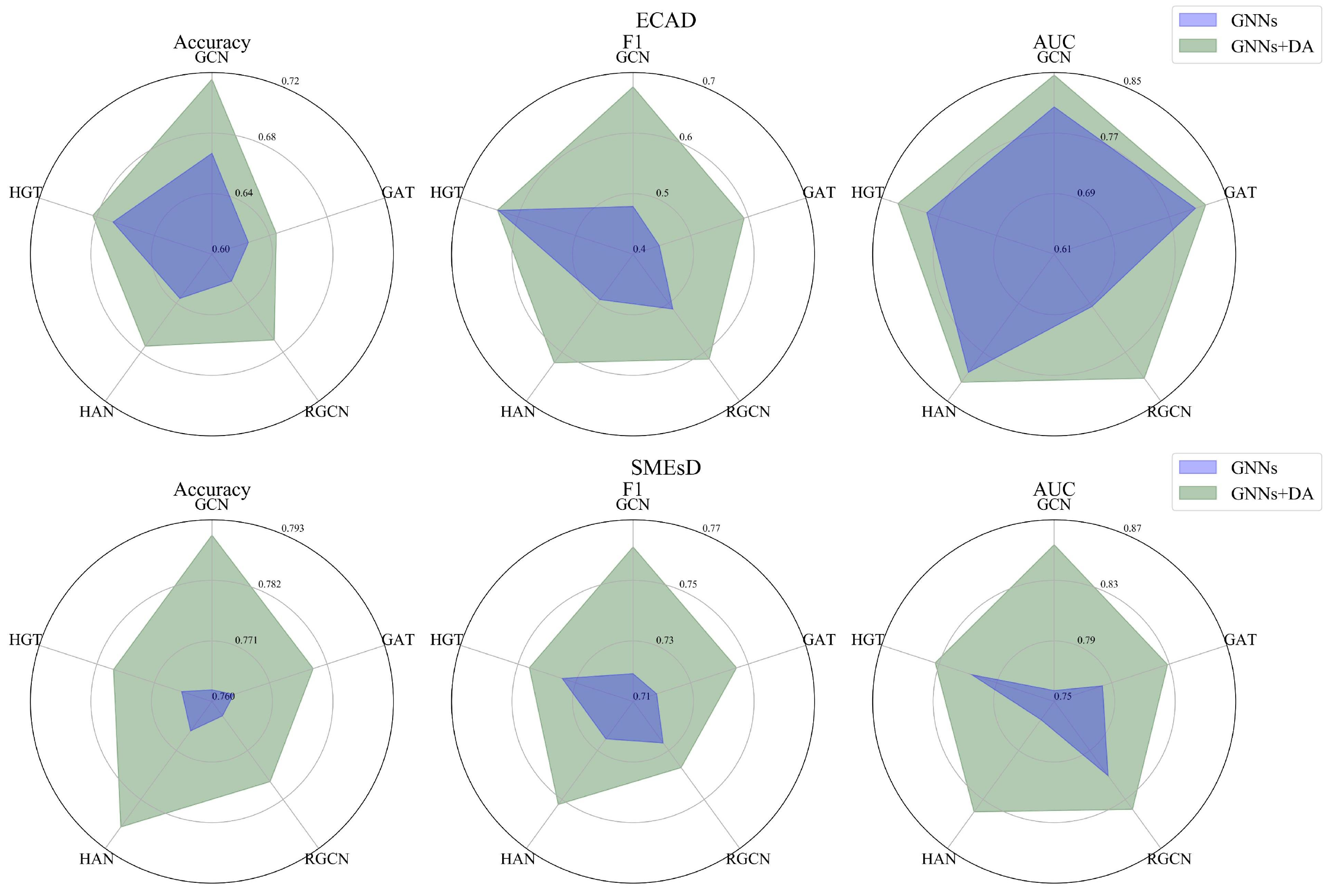}
    \caption{Dimension Attention Boosts GNNs. Experiments are conducted on ECAD dataset (upper part) and SMEsD dataset (lower part).
    The purple and green lines in the figures represent the performance of different base models and their augmented performance with dimension attention, evaluated across various comprehensive indicators.}
    \vspace{-1em}
    \label{fig:plug_in}
\end{figure*}

\subsection{Visualization Analysis}
We make an intuitive comparison by projecting enterprise representations into two dimensions using t-SNE \cite{van2008visualizing}. We show the visualization results in Figure \ref{fig:Visualization}, where different colors denote different ground truth risk levels. We can observe that the node representations generated by GDAN can be distinguished more easily compared to the other models (i.e., there is less mixing of the classes). Specifically, the enterprises in the upper area mostly belong to the class ``AAA'', the ones in the middle part mostly belong to the class ``AA'', and the ones in the left side mostly belong to ``Others''. 

\begin{figure}[h]
    \centering
    \includegraphics[width=\linewidth]
    {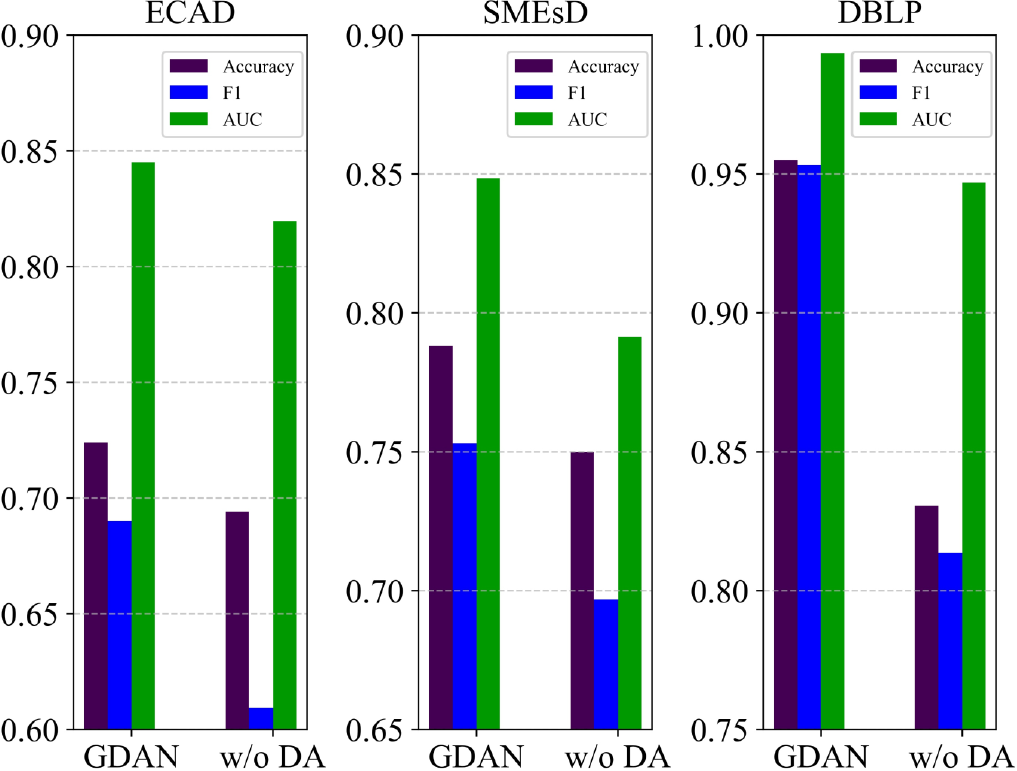}
    \caption{Ablation Study}
    \vspace{-1em}
    \label{fig:ablation_study}
\end{figure}

\subsection{Ablation Study}
To demonstrate the effectiveness of the dimension attention mechanism, we conduct an ablation study. We compare the performance of GDAN with an identical implementation without dimension attention. We can observe from Figure \ref{fig:ablation_study} that, compared with the model without dimension attention, GDAN does better in all indicators on the three datasets. The results confirm the notion that dimension-based attention can enhance the performance of traditional GNNs.

\subsection{Dimension Attention as Plug-in Module}
We demonstrate that the dimension attention module can serve as a plug-in module to augment various GNNs. Specifically, we integrate this module into popular GNNs such as GCN, GAT, RGCN \cite{Schlichtkrull2018Modeling}, HAN, and HGT, treating them as base models. We show the performances of GNNs and the augmented versions regarding the two financial risk datasets (i.e., ECAD and SMEsD) and three indicators (i.e., Accuracy, F1 and AUC). As depicted in Figure \ref{fig:plug_in}, it is evident that all the augmented base models consistently outperform their original counterparts.

\subsection{Parameter Analysis}
\begin{figure}[h]
    \centering
    \includegraphics[width=\linewidth]{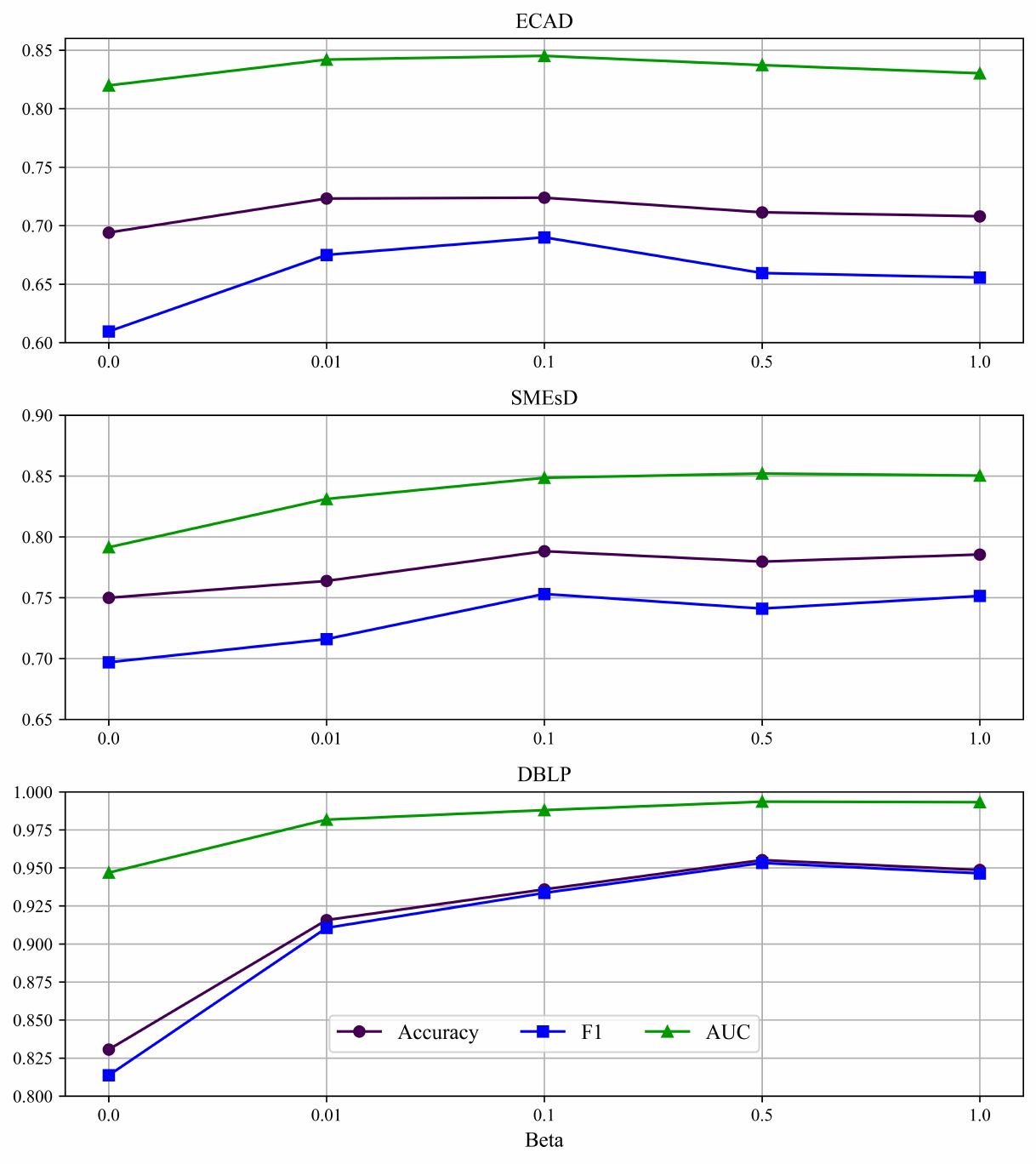}
    \caption{Parameter Analysis}
    \vspace{-1em}
    \label{fig:parameter_analysis}
\end{figure}

The balance between graph convolution-based representations and dimension attention-based representations is important for the final capability. We show the performance of GDAN for different values of $\beta$ on the DBLP dataset, ranging from $\{0,0.01,0.1,0.5,1\}$. We can observe from Figure \ref{fig:parameter_analysis} that a good $\beta$ value helps the model achieve better performance. When comparing the performance under the setting $\beta=0$ and $\beta=0.01$, we find that even a very small $\beta$, which means limited information from dimension-attention-based representations, can enhance the capacity of GNNs. 
On the other end, a large $\beta$ can lead to over-fitting, thus hurting the performance. 


\subsection{Interpretability}
\subsubsection{Intuition}
As the topology information in EHG provides a large gain for modeling credit risk level, we believe that finding important edges can help understanding contagion risk in EHG. The intuition of the experiment is that if an edge is important, then the edge should contribute to the performance of a GNN model. Thus, reserving important edges can help maintain the performance, and less important edges can be deleted as a result of limited contribution. On the contrary, deleting important edges should lead to more performance decrease. 
As illustrated in Table \ref{tab:gnn explainer method comparison}, we compare our proposed method with current representative techniques. Notably, we were unable to identify comparable approaches for DistShift. Most existing interpretability methods for GNNs focus on subgraphs as explanations. In contrast, we propose calculating edge-level importance for model explanation. Additionally, while these methods predominantly emphasize model behavior, our approach considers the data itself. Furthermore, most current methods rely on deep neural networks (DNNs) for interpretability. In the case of the data-centric method kNN-Shapley, node importance is determined in polynomial time relative to the number of nodes, requiring labeled data.
Thus, we compare the performances of the proposed DistShift with Leave One Out (LOO). LOO is a classical data evaluation method, which measures the importance of the target item by comparing the difference between deleting the target item and the whole dataset. 
We calculate edge importance using different methods. For the DistShift, we sample 10 subgraphs with 1 whole graph and average the importance for each edge to get the final stable edge importance.

\begin{table}[t]
\begin{center}
\caption{Comparison between DistShift and other GNN interpretability methods.}
\label{tab:gnn explainer method comparison}
\resizebox{0.49\textwidth}{!}{
\begin{tabular}{lccccc}
\toprule
\multicolumn{1}{c}{\multirow{2}{*}{Method}} & \multicolumn{2}{c}{Design} & \multicolumn{2}{c}{Technique} & \multicolumn{1}{c}{\multirow{2}{*}{Feasible}} \\
\cline{2-5}
\multicolumn{1}{c}{}                        & Edge-level  & Data-centric & Non-DNN   & Label irrelevant  & \multicolumn{1}{c}{}                          \\
\midrule
LOO                                         & \CheckmarkBold           & \CheckmarkBold            & \CheckmarkBold         & \CheckmarkBold                 & Yes                                           \\
kNN-Shapley                                 & \XSolidBrush          & \CheckmarkBold            & \CheckmarkBold         & \XSolidBrush                & No                                            \\
GNNExplainer  \cite{ying2019gnnexplainer}                              & \XSolidBrush          & \XSolidBrush           & \XSolidBrush        & \CheckmarkBold                 & No                                            \\
PGExplainer \cite{luo2020parameterized}                                & \XSolidBrush          & \XSolidBrush           & \XSolidBrush        & \CheckmarkBold                 & No                                            \\
ProtGNN  \cite{zhang2022protgnn}                                   & \XSolidBrush          & \XSolidBrush           & \XSolidBrush        & \CheckmarkBold                 & No                                            \\
SIGNET \cite{liu2024towards}                                     & \XSolidBrush          & \XSolidBrush           & \XSolidBrush        & \CheckmarkBold                 & No                                            \\
DistShift (Ours)                                   & \CheckmarkBold           & \CheckmarkBold            & \CheckmarkBold         & \CheckmarkBold                 &   - \\
\bottomrule
\end{tabular}
}
\end{center}
\end{table}

\begin{figure}[t]
    \centering
    
    \includegraphics[width=\linewidth]{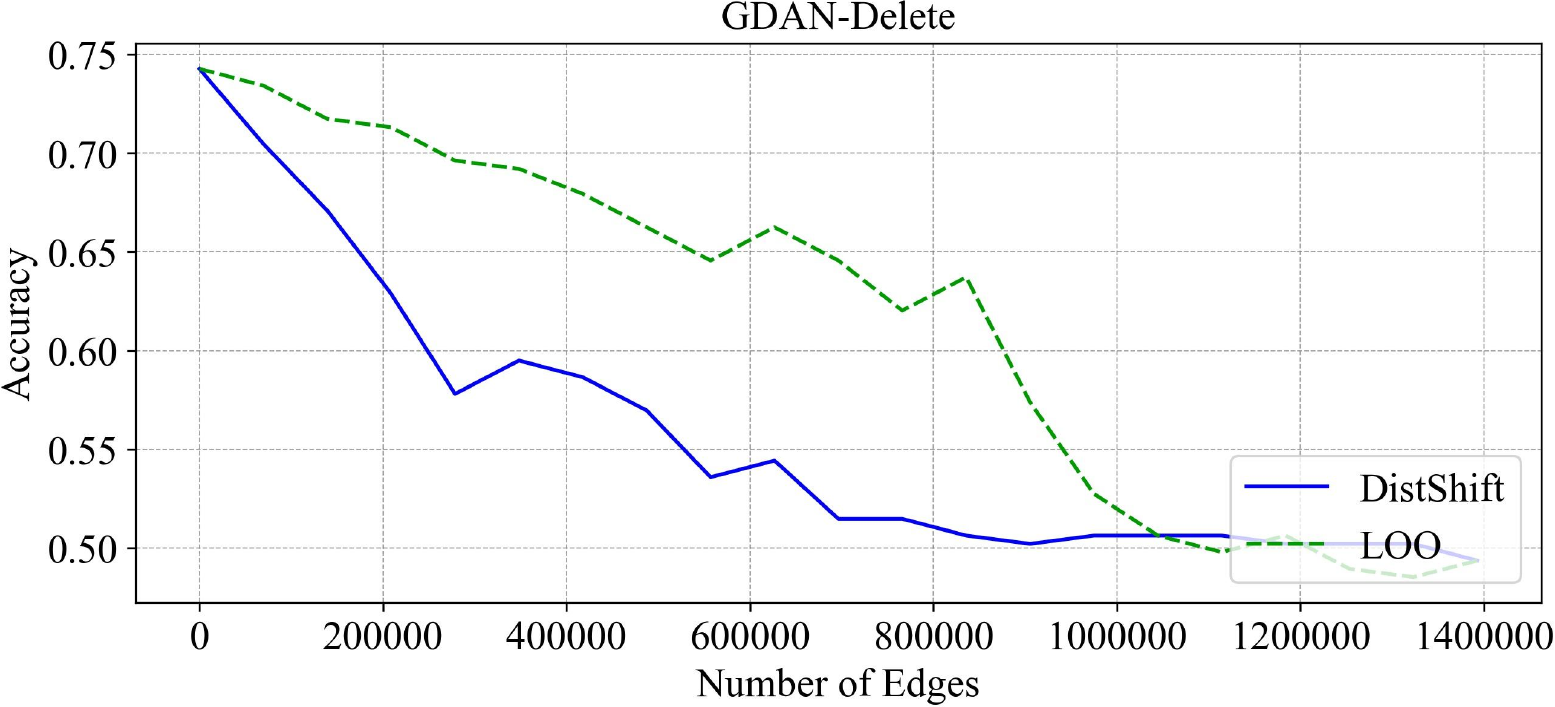}
    \includegraphics[width=\linewidth]{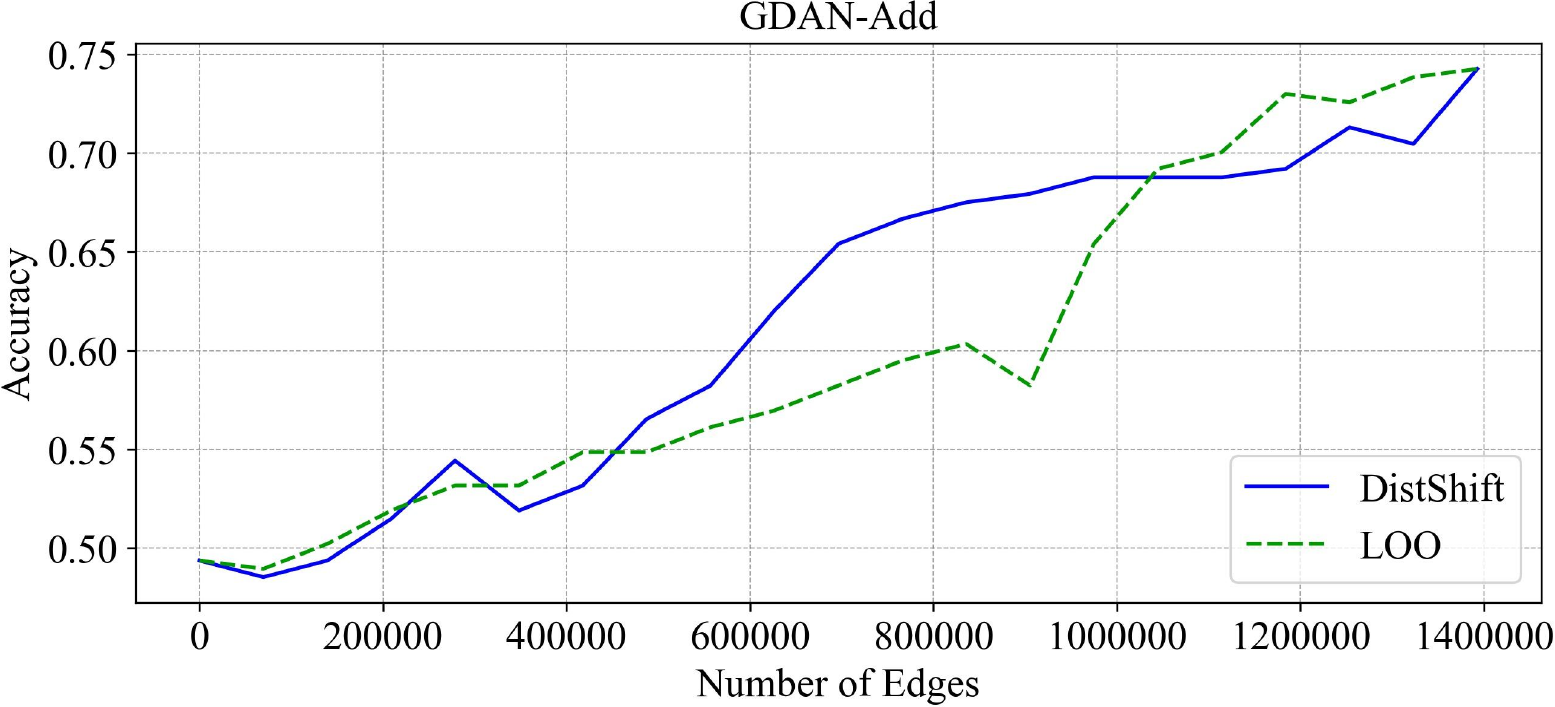}
    \caption{Edge-level Interpretability. 
    }
    \vspace{-1em}
    \label{fig:edge_importance}
\end{figure} 

\subsubsection{Results and Analysis}

We conduct experiments based on GDAN and ECAD dataset to evaluate the calculated importance of edges. 
In particular, we focus on two distinct scenarios: deleting and adding edges based on their varying importance. Initially, we started with the ECAD dataset containing all edges and systematically deleted edges in increments of 5\% of the total, prioritizing those with higher calculated importance. For each deletion, we retrained the GCN model from scratch to assess the impact. Analogously, in the addition scenario, we started with the ECAD dataset without any edges and systematically added edges in increments of 5\%, again prioritizing those with higher calculated importance.

From Figure \ref{fig:edge_importance}, we can see that the proposed DistShift outperforms LOO. 
In the GDAN-Delete subfigure, the DistShift curve is significantly lower than that of LOO, indicating that deleting the same number of edges results in a greater performance decrease. This further demonstrates that DistShift identifies the most critical edges. 
In addition, we can observe from the GDAN-Add figure that 
DistShift shows comparable performance to the other methods.
Notably, DistShift does not rely on deep neural networks that require more computing resources and offer less transparency.

\begin{figure}[t]
    \centering
    \includegraphics[width=\linewidth]{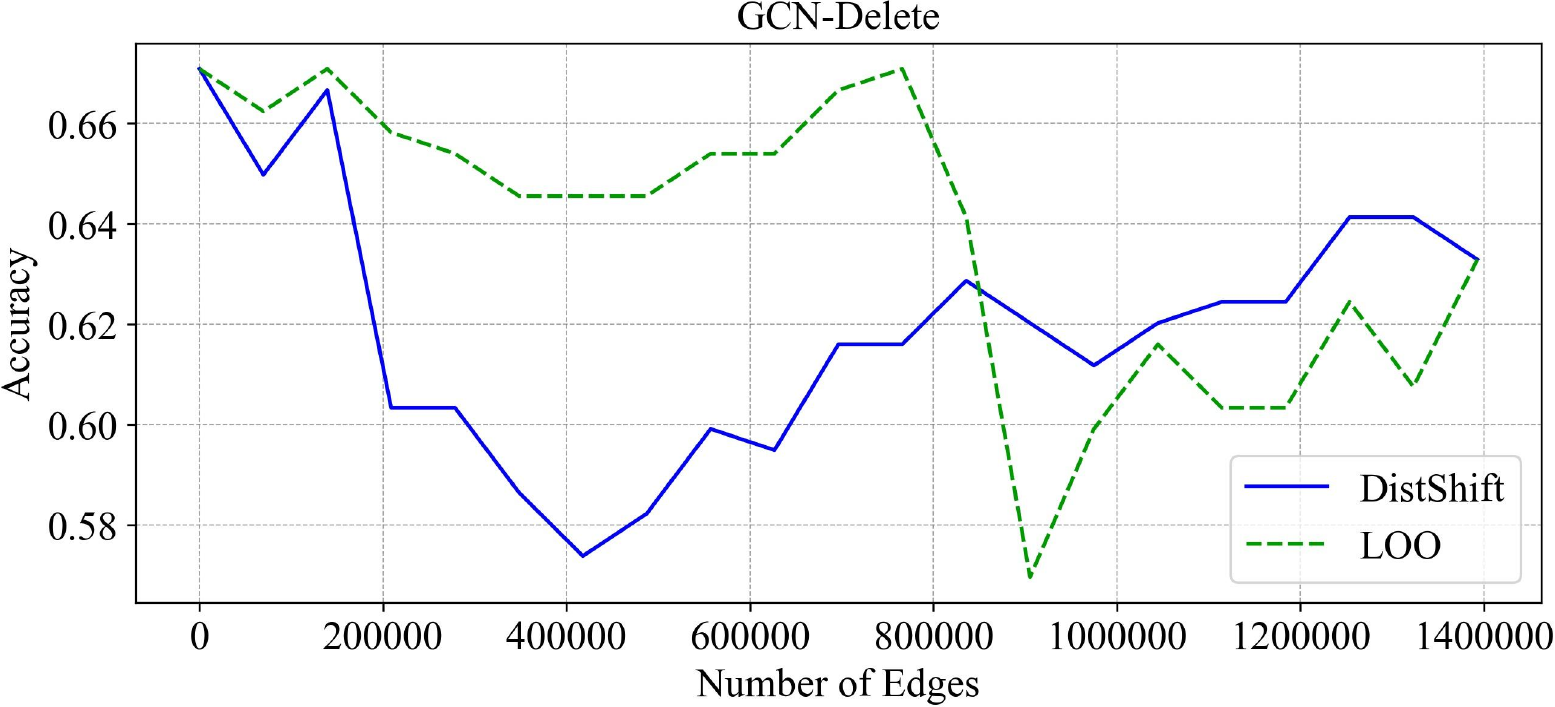}
    \includegraphics[width=\linewidth]{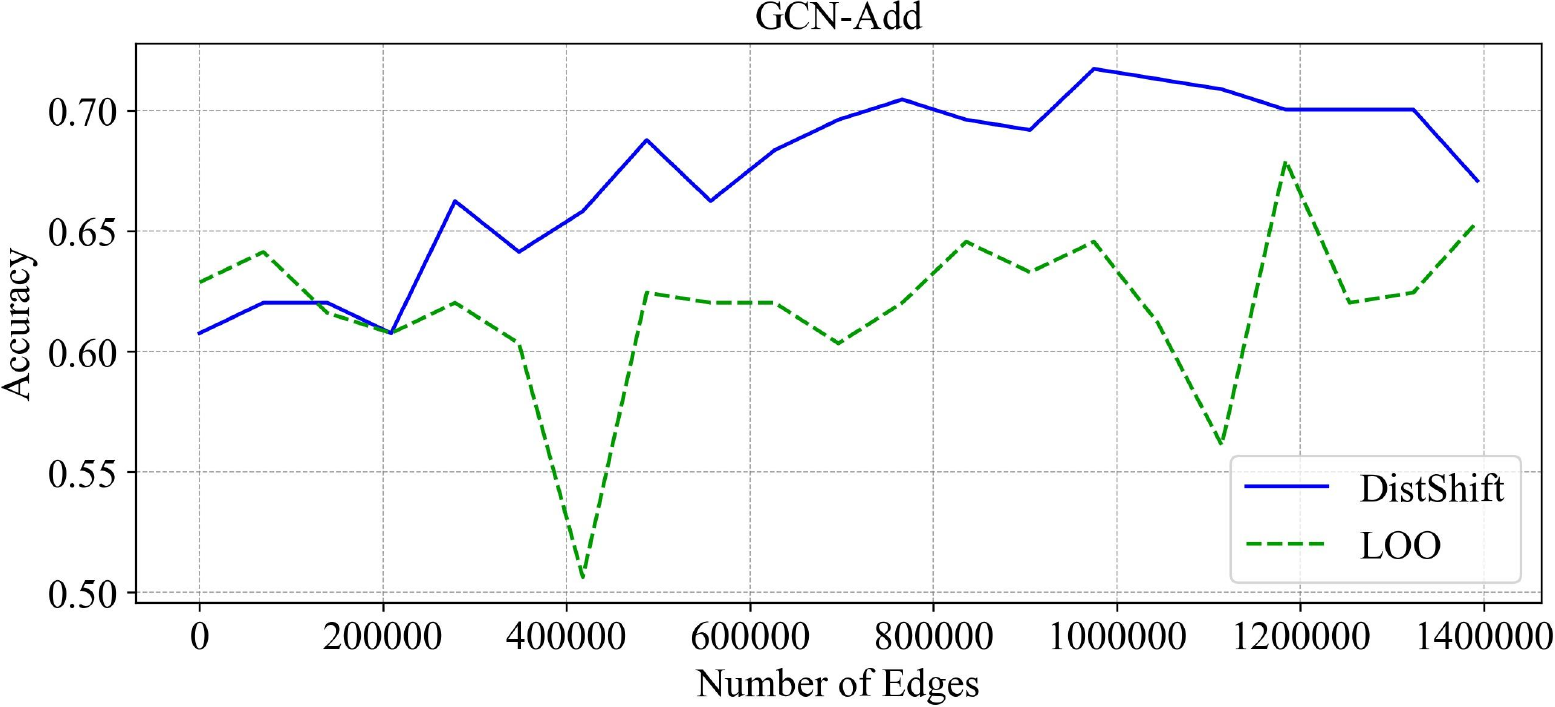}
    \caption{Supplementary Experiment Results. 
    }
    \vspace{-1em}
    \label{fig:Supplementary_Experiment}
\end{figure} 

\subsubsection{Supplementary Experiment}
 To comprehensively demonstrate the generalization capability of DistShift, we replicated these experiments using the GCN model.
 As illustrated in the GCN-Delete subfigure in Figure \ref{fig:Supplementary_Experiment}, DistShift show comparable performance.
In the GCN-Add subfigure, it is evident that we need only around 400,000 edges, or 30\% of the total edges, to achieve comparable performance with GCN. This indicates that DistShift effectively identifies the important edges that influence the distribution of features. Additionally, some edges negatively impact performance and should be removed during training, likely due to the presence of noisy edges in real-world data.


\section{Conclusion}
This study introduces an innovative approach to enterprise credit assessment through the application of Graph Dimension Attention Networks (GDAN), which can effectively discern subtle distinctions across various feature dimensions from the same neighboring node. This emphasis arises from the recognition that, even when considering the same neighbor, different feature dimensions may vary significantly in their relevance for assessing contagion risk. Given the paramount importance of interpretability in the financial context, we also propose GDAN-DistShift to provide edge-level interpretability for GDAN by identifying important edges in the message-passing process. 
In particular, DistShift aims to capture the influence that an edge has on a target node (i.e., whether an edge enhances the distinguishability of the node of interest).
This method employs an ML model as a surrogate, enhancing transparency in the explanation process.
To assess the performance of GDAN and DistShift, we assembled a multi-source real-world financial dataset, ECAD, and conducted a comprehensive series of experiments. We make this dataset publicly accessible to foster further exploration and advancement within this research domain.
In addition, we conduct experiments on publicly available datasets, SMEsD and DBLP. Comprehensive experimental results, including ablation studies, substantiate the effectiveness of the proposed methods. 
In the future, we will explore the dimension attention mechanism further, including its combination with other deep learning techniques and its application in other scenarios.

\section*{Acknowledgments}
The research is supported by the Key Technologies Research and Development Program under Grant No. 2020YFC0832702, 
and National Natural Science Foundation of China under Grant Nos. 71910107002, 62376227, 61906159, 62302400, 62176014, and Sichuan Science and Technology Program under Grant No. 2023NSFSC0032, 2023NSFSC0114, and Guanghua Talent Project of Southwestern University of Finance and Economics.

\bibliographystyle{IEEEtran}
\bibliography{sample-base}

\begin{thebibliography}{10}
\providecommand{\url}[1]{#1}
\csname url@samestyle\endcsname
\providecommand{\newblock}{\relax}
\providecommand{\bibinfo}[2]{#2}
\providecommand{\BIBentrySTDinterwordspacing}{\spaceskip=0pt\relax}
\providecommand{\BIBentryALTinterwordstretchfactor}{4}
\providecommand{\BIBentryALTinterwordspacing}{\spaceskip=\fontdimen2\font plus
\BIBentryALTinterwordstretchfactor\fontdimen3\font minus \fontdimen4\font\relax}
\providecommand{\BIBforeignlanguage}[2]{{%
\expandafter\ifx\csname l@#1\endcsname\relax
\typeout{** WARNING: IEEEtran.bst: No hyphenation pattern has been}%
\typeout{** loaded for the language `#1'. Using the pattern for}%
\typeout{** the default language instead.}%
\else
\language=\csname l@#1\endcsname
\fi
#2}}
\providecommand{\BIBdecl}{\relax}
\BIBdecl

\bibitem{jegadeesh1993returns}
N.~Jegadeesh and S.~Titman, ``Returns to buying winners and selling losers: Implications for stock market efficiency,'' \emph{The Journal of Finance}, vol.~48, no.~1, pp. 65--91, 1993.

\bibitem{yang2020financial}
S.~Yang, Z.~Zhang, J.~Zhou, Y.~Wang, W.~Sun, X.~Zhong, Y.~Fang, Q.~Yu, and Y.~Qi, ``Financial risk analysis for smes with graph-based supply chain mining,'' in \emph{Proceedings of IJCAI}, 2021, pp. 4661--4667.

\bibitem{wang2021temporal}
D.~Wang, Z.~Zhang, J.~Zhou, P.~Cui, J.~Fang, Q.~Jia, Y.~Fang, and Y.~Qi, ``Temporal-aware graph neural network for credit risk prediction,'' in \emph{Proceedings of SDM}.\hskip 1em plus 0.5em minus 0.4em\relax SIAM, 2021, pp. 702--710.

\bibitem{bi2022company}
W.~Bi, B.~Xu, X.~Sun, Z.~Wang, H.~Shen, and X.~Cheng, ``Company-as-tribe: Company financial risk assessment on tribe-style graph with hierarchical graph neural networks,'' in \emph{Proceedings of SIGKDD}, 2022, pp. 2712--2720.

\bibitem{zheng2023midlg}
W.~Zheng, B.~Xu, E.~Lu, Y.~Li, Q.~Cao, X.~Zong, and H.~Shen, ``Midlg: Mutual information based dual level gnn for transaction fraud complaint verification,'' in \emph{Proceedings of SIGKDD}, 2023, pp. 5685--5694.

\bibitem{lo1986logit}
A.~W. Lo, ``Logit versus discriminant analysis: A specification test and application to corporate bankruptcies,'' \emph{Journal of econometrics}, vol.~31, no.~2, pp. 151--178, 1986.

\bibitem{crouhy2000comparative}
M.~Crouhy, D.~Galai, and R.~Mark, ``A comparative analysis of current credit risk models,'' \emph{Journal of Banking \& Finance}, vol.~24, no. 1-2, pp. 59--117, 2000.

\bibitem{lopez2000evaluating}
J.~A. Lopez and M.~R. Saidenberg, ``Evaluating credit risk models,'' \emph{Journal of Banking \& Finance}, vol.~24, no. 1-2, pp. 151--165, 2000.

\bibitem{olson2012comparative}
D.~L. Olson, D.~Delen, and Y.~Meng, ``Comparative analysis of data mining methods for bankruptcy prediction,'' \emph{Decision Support Systems}, vol.~52, no.~2, pp. 464--473, 2012.

\bibitem{delen2013measuring}
D.~Delen, C.~Kuzey, and A.~Uyar, ``Measuring firm performance using financial ratios: A decision tree approach,'' \emph{Expert Systems with Applications}, vol.~40, no.~10, pp. 3970--3983, 2013.

\bibitem{tsai2008using}
C.-F. Tsai and J.-W. Wu, ``Using neural network ensembles for bankruptcy prediction and credit scoring,'' \emph{Expert systems with applications}, vol.~34, no.~4, pp. 2639--2649, 2008.

\bibitem{zhang2022credit}
W.~Zhang, S.~Yan, J.~Li, X.~Tian, and T.~Yoshida, ``Credit risk prediction of smes in supply chain finance by fusing demographic and behavioral data,'' \emph{Transportation Research Part E: Logistics and Transportation Review}, vol. 158, p. 102611, 2022.

\bibitem{craja2020deep}
P.~Craja, A.~Kim, and S.~Lessmann, ``Deep learning for detecting financial statement fraud,'' \emph{Decision Support Systems}, vol. 139, p. 113421, 2020.

\bibitem{borchert2023extending}
P.~Borchert, K.~Coussement, A.~De~Caigny, and J.~De~Weerdt, ``Extending business failure prediction models with textual website content using deep learning,'' \emph{European Journal of Operational Research}, vol. 306, no.~1, pp. 348--357, 2023.

\bibitem{chen2023bankruptcy}
T.-K. Chen, H.-H. Liao, G.-D. Chen, W.-H. Kang, and Y.-C. Lin, ``Bankruptcy prediction using machine learning models with the text-based communicative value of annual reports,'' \emph{Expert Systems with Applications}, vol. 233, p. 120714, 2023.

\bibitem{li2024corporate}
Q.~Li, H.~Shan, Y.~Tang, and V.~Yao, ``Corporate climate risk: Measurements and responses,'' \emph{The Review of Financial Studies}, 2024.

\bibitem{Rong2020Self}
Y.~Rong, Y.~Bian, T.~Xu, W.~Xie, Y.~Wei, W.~Huang, and J.~Huang, ``Self-supervised graph transformer on large-scale molecular data,'' in \emph{Proceedings of NIPS}, "2020".

\bibitem{hu2020heterogeneous}
Z.~Hu, Y.~Dong, K.~Wang, and Y.~Sun, ``Heterogeneous graph transformer,'' in \emph{Proceedings of WWW}, 2020, pp. 2704--2710.

\bibitem{huang2022contexting}
Y.-H. Huang, Y.-H. Chen, and Y.-S. Chen, ``{ConTextING}: Granting document-wise contextual embeddings to graph neural networks for inductive text classification,'' in \emph{Proceedings of ACL}, 2022, pp. 1163--1168.

\bibitem{xv2023commerce}
G.~Xv, C.~Lin, W.~Guan, J.~Gou, X.~Li, H.~Deng, J.~Xu, and B.~Zheng, ``E-commerce search via content collaborative graph neural network,'' in \emph{Proceedings of SIGKDD}, 2023, pp. 2885--2897.

\bibitem{Xu2017Scene}
D.~Xu, Y.~Zhu, C.~B. Choy, and L.~Fei-Fei, ``Scene graph generation by iterative message passing,'' in \emph{Proceedings of CVPR}, 2017.

\bibitem{Kipf2017Semi-supervised}
T.~N. Kipf and M.~Welling, ``Semi-supervised classification with graph convolutional networks,'' in \emph{Proceedings of ICLR}, 2017.

\bibitem{tsitsulin2023graph}
A.~Tsitsulin, J.~Palowitch, B.~Perozzi, and E.~M{\"u}ller, ``Graph clustering with graph neural networks,'' \emph{Journal of Machine Learning Research}, vol.~24, no. 127, pp. 1--21, 2023.

\bibitem{Wang2019Heterogeneous}
X.~Wang, H.~Ji, C.~Shi, B.~Wang, Y.~Ye, P.~Cui, and P.~S. Yu, ``Heterogeneous graph attention network,'' in \emph{Proceddings of WWW}, 2019, pp. 2022--2032.

\bibitem{fountoulakis2023graph}
K.~Fountoulakis, A.~Levi, S.~Yang, A.~Baranwal, and A.~Jagannath, ``Graph attention retrospective,'' \emph{JMLR}, vol.~24, no. 246, pp. 1--52, 2023.

\bibitem{Velickovic2018Graph}
P.~Veli{\v{c}}kovi{\'c}, G.~Cucurull, A.~Casanova, A.~Romero, P.~Li{\`o}, and Y.~Bengio, ``Graph attention networks,'' in \emph{Proceedings of ICLR}, 2018.

\bibitem{lee2023towards}
S.~Y. Lee, F.~Bu, J.~Yoo, and K.~Shin, ``Towards deep attention in graph neural networks: Problems and remedies,'' in \emph{Proceedings of ICML}.\hskip 1em plus 0.5em minus 0.4em\relax PMLR, 2023, pp. 18\,774--18\,795.

\bibitem{yang2021financial}
S.~Yang, Z.~Zhang, J.~Zhou, Y.~Wang, W.~Sun, X.~Zhong, Y.~Fang, Q.~Yu, and Y.~Qi, ``Financial risk analysis for smes with graph-based supply chain mining,'' in \emph{Proceedings of IJCAI}, 2021, pp. 4661--4667.

\bibitem{zheng2021heterogeneous}
Y.~Zheng, V.~C. Lee, Z.~Wu, and S.~Pan, ``Heterogeneous graph attention network for small and medium-sized enterprises bankruptcy prediction,'' in \emph{Proceedings of PAKDD}, 2021, pp. 140--151.

\bibitem{xiang2023semi}
S.~Xiang, M.~Zhu, D.~Cheng, E.~Li, R.~Zhao, Y.~Ouyang, L.~Chen, and Y.~Zheng, ``Semi-supervised credit card fraud detection via attribute-driven graph representation,'' in \emph{Proceedings of AAAI}, vol.~37, no.~12, 2023, pp. 14\,557--14\,565.

\bibitem{shi2024improved}
Y.~Shi, Y.~Qu, Z.~Chen, Y.~Mi, and Y.~Wang, ``Improved credit risk prediction based on an integrated graph representation learning approach with graph transformation,'' \emph{European Journal of Operational Research}, vol. 315, no.~2, pp. 786--801, 2024.

\bibitem{wei2024combining}
S.~Wei, J.~Lv, Y.~Guo, Q.~Yang, X.~Chen, Y.~Zhao, Q.~Li, F.~Zhuang, and G.~Kou, ``Combining intra-risk and contagion risk for enterprise bankruptcy prediction using graph neural networks,'' \emph{Information Sciences}, p. 120081, 2024.

\bibitem{nagarajan2019uniform}
V.~Nagarajan and J.~Z. Kolter, ``Uniform convergence may be unable to explain generalization in deep learning,'' in \emph{Proceedings of NeurIPS}, vol.~32, 2019.

\bibitem{angelov2020towards}
P.~Angelov and E.~Soares, ``Towards explainable deep neural networks (xdnn),'' \emph{Neural Networks}, vol. 130, pp. 185--194, 2020.

\bibitem{liu2021mining}
Q.~Liu, Z.~Liu, H.~Zhang, Y.~Chen, and J.~Zhu, ``Mining cross features for financial credit risk assessment,'' in \emph{Proceedings of CIKM}, 2021, pp. 1069--1078.

\bibitem{dumitrescu2022machine}
E.~Dumitrescu, S.~Hu{\'e}, C.~Hurlin, and S.~Tokpavi, ``Machine learning for credit scoring: Improving logistic regression with non-linear decision-tree effects,'' \emph{European Journal of Operational Research}, vol. 297, no.~3, pp. 1178--1192, 2022.

\bibitem{ying2019gnnexplainer}
Z.~Ying, D.~Bourgeois, J.~You, M.~Zitnik, and J.~Leskovec, ``Gnnexplainer: Generating explanations for graph neural networks,'' in \emph{Proceddings of NeurIPS}, vol.~32, 2019.

\bibitem{yuan2021explainability}
H.~Yuan, H.~Yu, J.~Wang, K.~Li, and S.~Ji, ``On explainability of graph neural networks via subgraph explorations,'' in \emph{Proceedings of ICML}.\hskip 1em plus 0.5em minus 0.4em\relax PMLR, 2021, pp. 12\,241--12\,252.

\bibitem{yin2023train}
J.~Yin, C.~Li, H.~Yan, J.~Lian, and S.~Wang, ``Train once and explain everywhere: Pre-training interpretable graph neural networks,'' in \emph{Proceedings of NeurIPS}, vol.~36, 2023.

\bibitem{chen2020revisiting}
L.~Chen, L.~Wu, R.~Hong, K.~Zhang, and M.~Wang, ``Revisiting graph based collaborative filtering: A linear residual graph convolutional network approach,'' in \emph{Proceedings of AAAI}, vol.~34, no.~01, 2020, pp. 27--34.

\bibitem{he2020lightgcn}
X.~He, K.~Deng, X.~Wang, Y.~Li, Y.~Zhang, and M.~Wang, ``Lightgcn: Simplifying and powering graph convolution network for recommendation,'' in \emph{Proceedings of SIGIR}, 2020, pp. 639--648.

\bibitem{wu2019simplifying}
F.~Wu, A.~Souza, T.~Zhang, C.~Fifty, T.~Yu, and K.~Weinberger, ``Simplifying graph convolutional networks,'' in \emph{Proceedings of ICML}.\hskip 1em plus 0.5em minus 0.4em\relax PMLR, 2019, pp. 6861--6871.

\bibitem{fey2019fast}
M.~Fey and J.~E. Lenssen, ``Fast graph representation learning with pytorch geometric,'' \emph{arXiv preprint arXiv:1903.02428}, 2019.

\bibitem{hosmer2013applied}
D.~W. Hosmer~Jr, S.~Lemeshow, and R.~X. Sturdivant, \emph{Applied logistic regression}.\hskip 1em plus 0.5em minus 0.4em\relax John Wiley \& Sons, 2013, vol. 398.

\bibitem{suykens1999least}
J.~A. Suykens and J.~Vandewalle, ``Least squares support vector machine classifiers,'' \emph{Neural Processing Letters}, vol.~9, pp. 293--300, 1999.

\bibitem{friedman2001greedy}
J.~H. Friedman, ``Greedy function approximation: a gradient boosting machine,'' \emph{Annals of Statistics}, pp. 1189--1232, 2001.

\bibitem{yang2023simple}
X.~Yang, M.~Yan, S.~Pan, X.~Ye, and D.~Fan, ``Simple and efficient heterogeneous graph neural network,'' in \emph{Proceedings of AAAI}, vol.~37, no.~9, 2023, pp. 10\,816--10\,824.

\bibitem{loshchilov2017decoupled}
I.~Loshchilov and F.~Hutter, ``Decoupled weight decay regularization,'' \emph{arXiv preprint arXiv:1711.05101}, 2017.

\bibitem{loshchilov2016sgdr}
{I. Loshchilov and F. Hutter}, ``{SGDR}: Stochastic gradient descent with warm restarts,'' in \emph{Proceddings of ICLR}, 2017.

\bibitem{van2008visualizing}
L.~Van~der Maaten and G.~Hinton, ``Visualizing data using t-sne.'' \emph{JMLR}, vol.~9, no.~11, 2008.

\bibitem{Schlichtkrull2018Modeling}
M.~Schlichtkrull, T.~N. Kipf, P.~Bloem, R.~van~den Berg, I.~Titov, and M.~Welling, ``Modeling relational data with graph convolutional networks,'' in \emph{Proceddings of ESWC}.\hskip 1em plus 0.5em minus 0.4em\relax Springer, 2018, pp. 593--607.

\bibitem{luo2020parameterized}
D.~Luo, W.~Cheng, D.~Xu, W.~Yu, B.~Zong, H.~Chen, and X.~Zhang, ``Parameterized explainer for graph neural network,'' in \emph{Proceedings of NeurIPS}, vol.~33, 2020, pp. 19\,620--19\,631.

\bibitem{zhang2022protgnn}
Z.~Zhang, Q.~Liu, H.~Wang, C.~Lu, and C.~Lee, ``Protgnn: Towards self-explaining graph neural networks,'' in \emph{Proceedings of AAAI}, vol.~36, no.~8, 2022, pp. 9127--9135.

\bibitem{liu2024towards}
Y.~Liu, K.~Ding, Q.~Lu, F.~Li, L.~Y. Zhang, and S.~Pan, ``Towards self-interpretable graph-level anomaly detection,'' in \emph{Proceedings of NeurIPS}, vol.~36, 2024.

\end{thebibliography}

\begin{IEEEbiography}[{\includegraphics[width=1in,height=1.25in,clip,keepaspectratio]{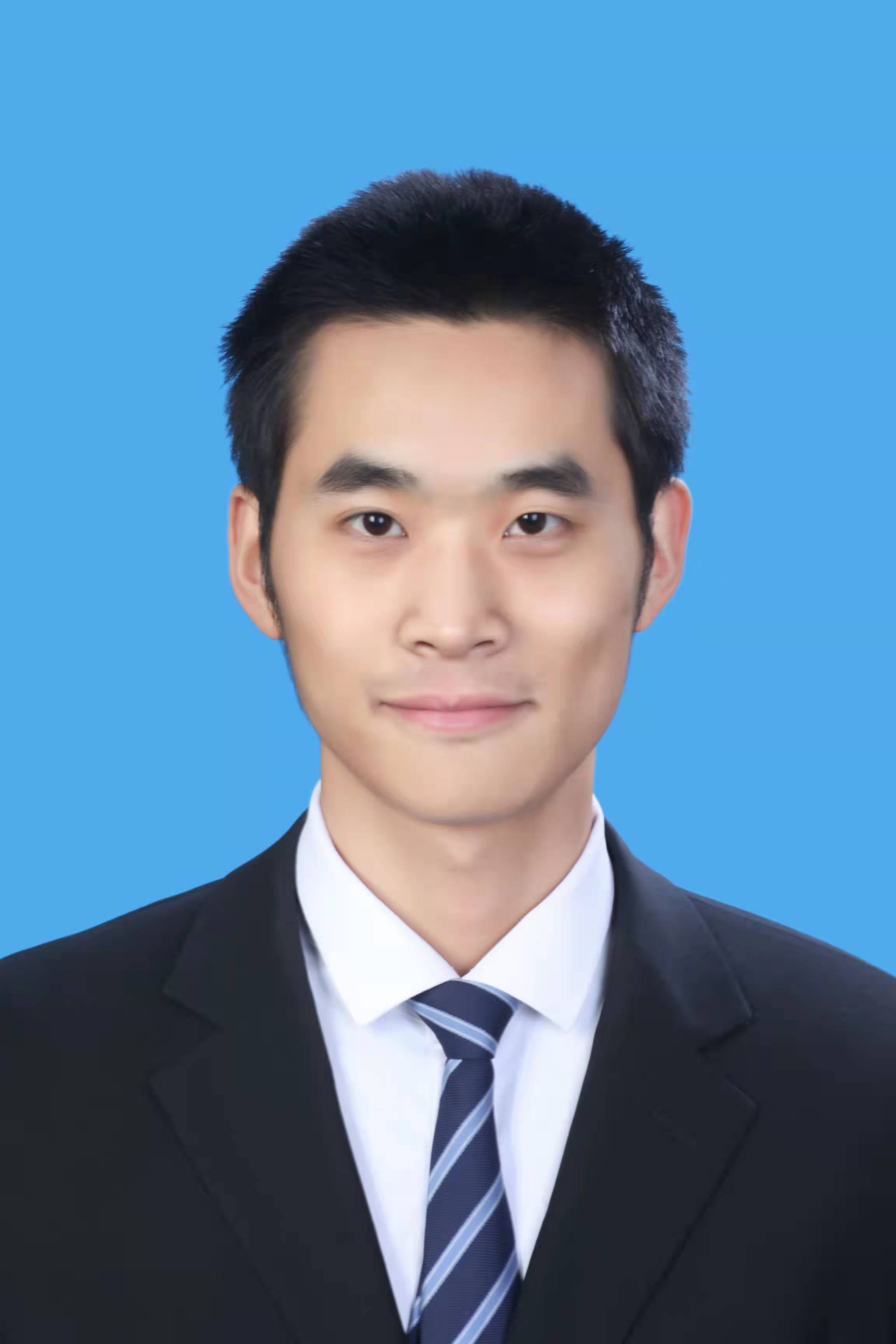}}]{Shaopeng Wei} received the B.S. degree from Huazhong Agricultural University in 2019, and Ph.D. degree from Southwestern University of Finance and Economics in 2024. He is currently an Assistant Professor with Guangxi University.
His research interests include graph learning and relevant applications in Fintech. He has published papers on top journals, such as IEEE TKDE.
\end{IEEEbiography}

\begin{IEEEbiography}[{\includegraphics[width=1in,height=1.25in,clip,keepaspectratio]{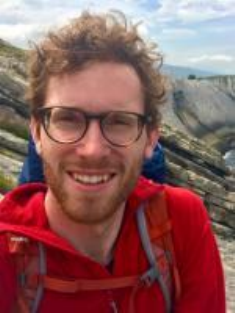}}]{Béni Egressy} received the B.S. degree from University of Cambridge in 2014, and now is a Ph.D. student in ETH Zürich. His research interests include graph neural networks and Fintech. He has published papers in NeurIPS, AAAI, EMNLP, etc.
\end{IEEEbiography}

\begin{IEEEbiography}[{\includegraphics[width=1in,height=1.25in,clip,keepaspectratio]{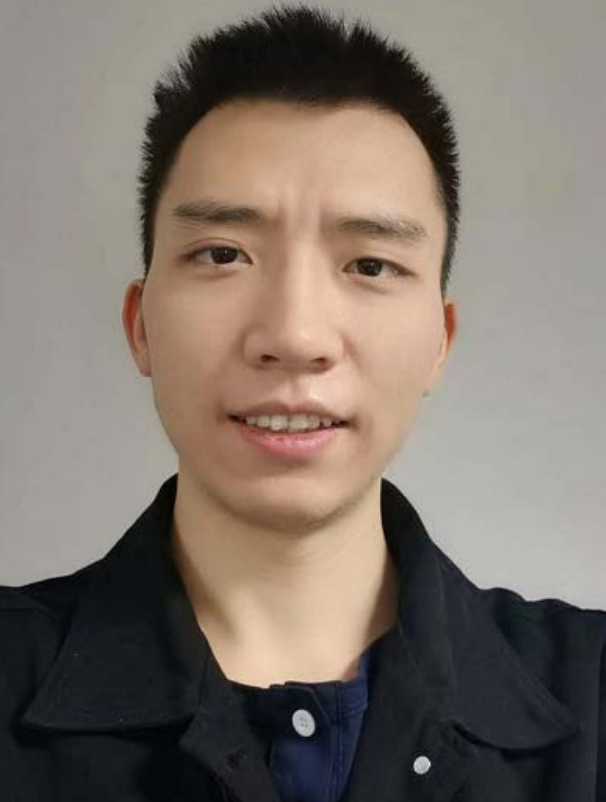}}]{Xingyan Chen}
	received the Ph.D. degree in computer technology from Beijing University of Posts and Telecommunications, in 2021. 
	He is currently an Associate Professor with Southwestern University of Finance and Economics.
	He has published papers on IEEE TKDE, \textsc{IEEE TMC}, \textsc{IEEE TCSVT}, \textsc{IEEE TII}, and \textsc{IEEE INFOCOM} etc. 
	His research interests include multimedia communications, multi-agent reinforcement learning and stochastic optimization.
\end{IEEEbiography}

\begin{IEEEbiography}[{\includegraphics[width=1in,height=1.25in,clip,keepaspectratio]{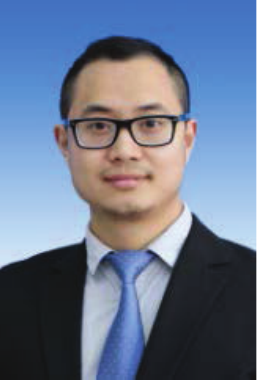}}]{Yu Zhao}
received the B.S. degree from Southwest Jiaotong University in 2006, and Ph.D. degrees from the Beijing University of Posts and Telecommunications in 2017. He is currently a full Professor at Southwestern University of Finance and Economics. His current research interests include machine learning, NLP, knowledge graph, Fintech. He has authored more than 30 papers in top journals and conferences including IEEE TKDE, IEEE TNNLS, IEEE TMC, KDD, ACL, ICME, etc.
\end{IEEEbiography}

\begin{IEEEbiography}[{\includegraphics[width=1in,height=1.25in,clip,keepaspectratio]{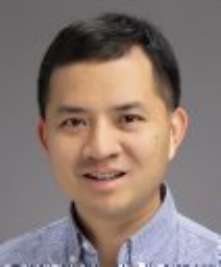}}]{Fuzhen Zhuang} received the Ph.D. degree in computer science from the Institute of Computing Technology, Chinese Academy of Sciences. 
He is currently a full Professor in Institute of Artificial Intelligence, Beihang University, China. 
His research interests include Machine Learning and Data Mining, including transfer learning, multi-task learning, multi-view learning and recommendation systems. 
He has published more than 100 papers in the prestigious refereed conferences and journals, such as KDD, WWW, SIGIR, ICDE, IJCAI, AAAI, EMNLP, Nature Communications, IEEE TKDE, ACM TKDD, IEEE T-CYB, IEEE TNNLS, ACM TIST, etc.
\end{IEEEbiography}

\begin{IEEEbiography}[{\includegraphics[width=1in,height=1.25in,clip,keepaspectratio]{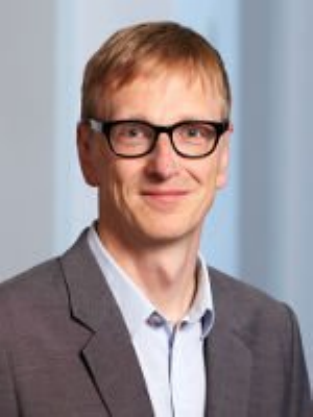}}]{Roger Wattenhofer} is a full professor at ETH Zürich, Switzerland. He received his doctorate in Computer Science from ETH Zürich. 
Roger Wattenhofer’s research interests include a variety of algorithmic and systems aspects in computer science and information technology.
He publishes in different communities: distributed computing (e.g., PODC, SPAA, DISC), networking and systems (e.g., SIGCOMM, SenSys, IPSN, OSDI, MobiCom), algorithmic theory (e.g., STOC, FOCS, SODA, ICALP), and more recently also machine learning (e.g., ICML, NeurIPS, ICLR, ACL, AAAI).
\end{IEEEbiography}

\begin{IEEEbiography}[{\includegraphics[width=1in,height=1.25in,clip,keepaspectratio]{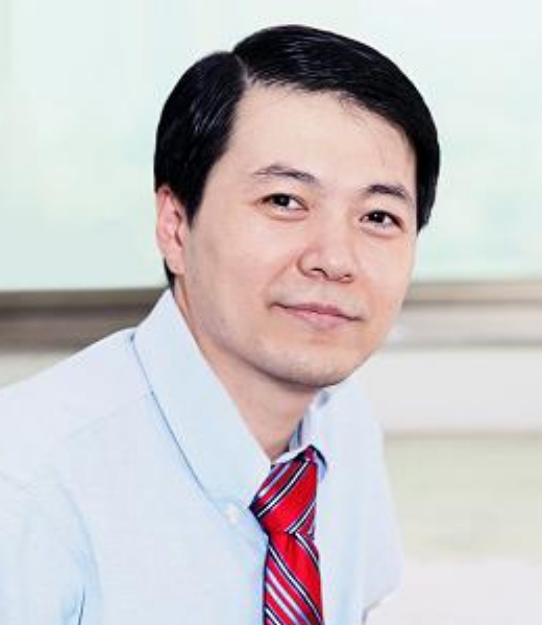}}]{Gang Kou} is a Distinguished Professor of Chang Jiang Scholars Program in Southwestern University of Finance and Economics.
He received his Ph.D. in Univ. of Nebraska at Omaha and B.S. degree in Tsinghua University, China. 
He has published more than 100 papers in various peer-reviewed journals. Gang Kou’s h-index is 72 and his papers have been cited for more than 19000 times. He is listed as the Highly Cited Researcher by Clarivate Analytics (Web of Science).
\end{IEEEbiography}







\end{document}